\documentclass{article}

     \PassOptionsToPackage{numbers, compress}{natbib}

\usepackage[preprint]{neurips_2025}

\usepackage[utf8]{inputenc} 
\usepackage[T1]{fontenc}    
\usepackage{hyperref}       
\usepackage{url}            
\usepackage{booktabs}       
\usepackage{amsfonts}       
\usepackage{nicefrac}       
\usepackage{microtype}      
\usepackage{xcolor}         
\usepackage{amsmath}
\usepackage{amssymb}
\usepackage{amsthm}
\usepackage{physics}
\usepackage{graphicx}
\usepackage{float}
\usepackage{caption}
\usepackage{subcaption}
\usepackage{multirow}
\usepackage{algorithm}
\usepackage{algpseudocode}

\theoremstyle{plain}
\newtheorem{theorem}{Theorem}[section]

\theoremstyle{definition}

\theoremstyle{remark}
\newtheorem{remark}[theorem]{Remark}

\title{GeoMaNO: Geometric Mamba Neural Operator \\ for Partial Differential Equations}

\newcommand*\samethanks[1][\value{footnote}]{\footnotemark[#1]}

\author{%
  Xi Han\thanks{These authors contributed equally to this paper.} \\ 
  Department of Computer Science \\
  Stony Brook University \\ 
  Stony Brook, NY 11794, USA \\
  \texttt{xihan1@cs.stonybrook.edu}
  \And
  Jingwei Zhang\samethanks[1] \\
  Department of Computer Science \\
  Stony Brook University \\ 
  Stony Brook, NY 11794, USA \\
  \texttt{jzhang@cs.stonybrook.edu} \\
  \And 
  Dimitris Samaras \\ 
  Department of Computer Science \\
  Stony Brook University \\ 
  Stony Brook, NY 11794, USA \\
  \texttt{samaras@cs.stonybrook.edu} \\
  \And
  Fei Hou \\ 
  Institute of Software \\
  Chinese Academy of Sciences \\
  Beijing, 100190, China \\ 
  \texttt{houfei@ios.ac.cn}
  \And 
  Hong Qin \\ 
  Department of Computer Science \\
  Stony Brook University \\ 
  Stony Brook, NY 11794, USA \\
  \texttt{qin@cs.stonybrook.edu}
}

\begin{document}

\maketitle

\begin{abstract}
The neural operator (NO) framework has emerged as a powerful tool for solving partial differential equations (PDEs). Recent NOs are dominated by the Transformer architecture, which offers NOs the capability to capture long-range dependencies in PDE dynamics. However, existing Transformer-based NOs suffer from quadratic complexity, lack geometric rigor, and thus suffer from sub-optimal performance on regular grids. As a remedy, we propose the Geometric Mamba Neural Operator (GeoMaNO) framework, which empowers NOs with Mamba's modeling capability, linear complexity, plus geometric rigor. We evaluate GeoMaNO's performance on multiple standard and popularly employed PDE benchmarks, spanning from Darcy flow problems to Navier-Stokes problems. GeoMaNO improves existing baselines in solution operator approximation by as much as $58.9\%$. 
\end{abstract}

\section{Introduction}

\paragraph{Background and Major Challenges} Partial differential equations (PDEs) are quintessential to various computational problems in science, engineering, as well as relevant applications in simulation, modeling, and scientific computing. Neural operators (NOs) have emerged as a powerful paradigm for learning mappings between function spaces. By learning the solution operator to PDEs, NOs offer a novel data-driven alternative to classical numerical solvers. 

Recent advances in the field of NOs have been propelled by the Transformer architecture~\cite{vaswani17-nips-transformer} for its modeling capability in capturing long-range dependencies and physical relations. However, Transformer-based NOs suffer from quadratic computational complexity inherent to the self-attention mechanism. Their architecture also lacks geometric rigor, and struggles when applied to large-scale, high-dimensional PDEs on regular grids. These shortcomings incur prohibitive computational costs, significantly degrading the inference accuracy and efficiency~\cite{hao23-icml-gnot, wu24-icml-transolver}. 

On the other hand, the Mamba architecture~\cite{gu23-colm-mamba, gu24-icml-mamba2}, benefiting from Transformer's parallelism and modeling capability, yet with a linear time complexity, has emerged as a promising avenue for neural operator instantiation. Research work like \cite{zheng24-nips-aliasfreemambaneuraloperator, hu25-corr-ssms-are-nos} tried to integrate Mamba with NOs, yet, these methods still lack geometric rigor on regular grids, and suffer from sub-optimal performance. Mamba's potential in scientific computation remains largely unexplored. 

\begin{figure}[ht]
    \centering
    \includegraphics[width=\linewidth]{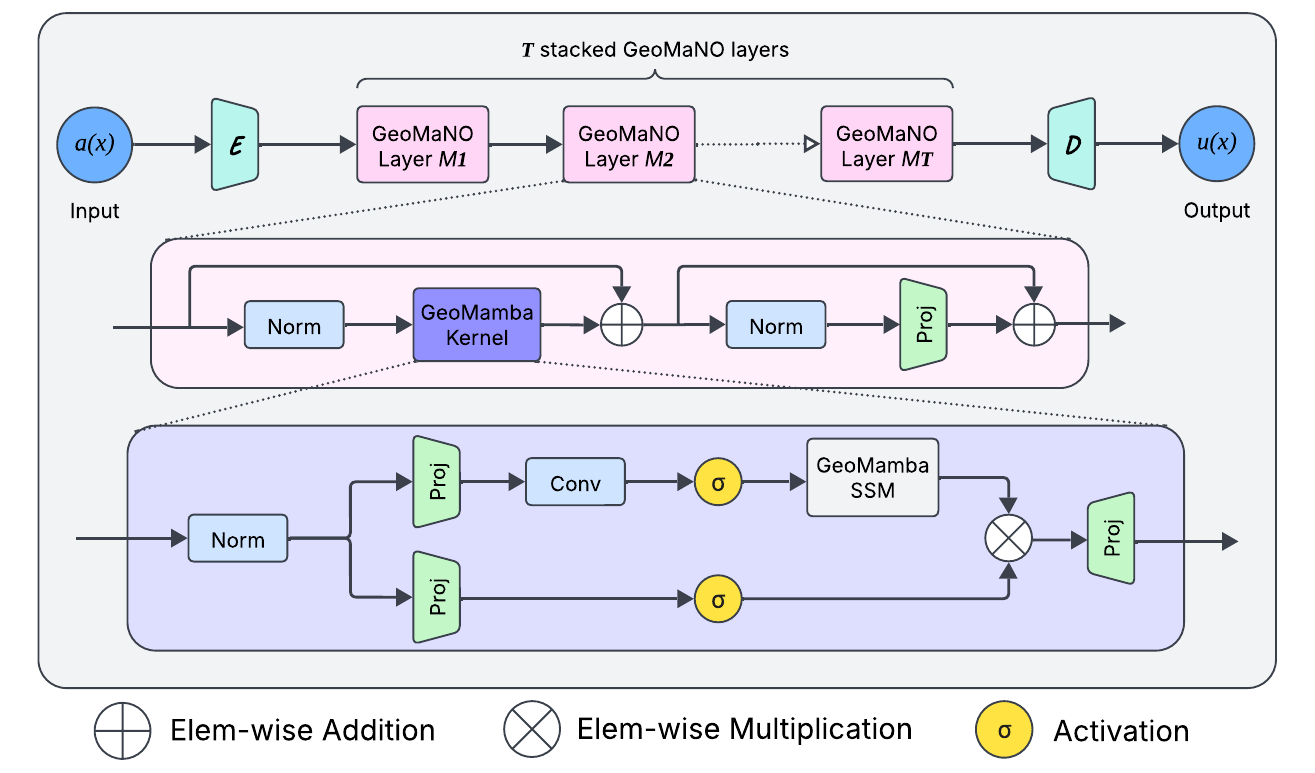}
    \vspace{-10pt}
    \caption[The novel GeoMaNO architecture.]{\textbf{Overview of the novel GeoMaNO architecture}. (1) The input function $a(x)$~\footnotemark~is lifted and patchified to a higher-dimensional latent representation by the geometry-aware encoder layer $\mathcal{E}$; (2) $T$ stacked GeoMaNO layers $M_1 \cdots M_T$ (middle) performs the kernel integral via GeoMamba kernels (bottom); (3) Each GeoMamba kernel performs geometry-aware Mamba scans, combined with skip-connections and non-linear activations; and (4) The last GeoMaNO layer's output is transformed back to the physical domain with the decoder layer $\mathcal{D}$. This yields the output function $u(x)$.}
    \label{fig:geomano-arch}
\end{figure}

\paragraph{Motivation and Method Overview} As a remedy to the challenges detailed above, we propose our novel \textbf{\textit{Geo}}metric \textbf{\textit{Ma}}mba \textbf{\textit{N}}eural \textbf{\textit{O}}perator (\textbf{\textit{GeoMaNO}}) framework, which is illustrated in Fig.~\ref{fig:geomano-arch}. GeoMaNO manifests a geometric-aware architecture specifically tailored for regular grids, combined with our GeoMamba-SSM module with geometric rigor and mathematical insights. GeoMamba-SSM employs a geometric correction component to the SSM formulation, effectively dampening duplicate hidden states introduced by multi-way cross-scans~\cite{liu24-nips-vmamba, zhu24-icml-vim}. For 1D and 3D problems, GeoMamba-SSM employs a one-dimensional geometry-aware state-space representation; to further enhance its geometric rigor, for 2D problems, GeoMamba-SSM employs a two-dimensional state-space representation, as inspired by~\cite{zhang25-cvpr-2dmamba}. 

\footnotetext{E.g., for the Darcy flow problem $\nabla \cdot (f(x) \nabla u(x)) = g(x)$, $a(x)$ encodes $f$ and $g$.}

\paragraph{Key Contributions} The salient contributions of this paper comprise: 
(1) \textbf{\textit{Theoretical insights}}: We propose a novel GeoMamba-SSM formulation, which manifests a geometrically rigorous state-space representation. It preserves the geometry of the input domain, and employs a learnable geometric correction component to dampen duplicate hidden states arising from multi-way cross-scans~\cite{liu24-nips-vmamba, zhu24-icml-vim}. 
(2) \textbf{\textit{The novel GeoMaNO architecture}}: We propose a novel family of neural operators, GeoMaNO, which utilizes GeoMamba-SSM's geometric rigor to approximate PDE solution operators, offering users high efficiency, high accuracy, high scalability, and high generalization power; and
(3) \textbf{\textit{Superior performance}}: We evaluate GeoMaNO's performance on the standard PDE solving benchmark~\cite{thuml25-online-pde-solving-std-benchmark}, which is popularly empolyed by NO frameworks. GeoMaNO improves the SOTA by $58.9\%$ (accuracy) and $29.2\%$ (efficiency) on the Navier-Stokes benchmark, and $36.8\%$ (accuracy) and $49.8\%$ (efficiency) on the Darcy flow benchmark.

\section{Related Work}

\paragraph{Neural Operators} The NO framework learns mappings between function spaces. The foundational work, DeepONet~\cite{lu21-nature-deeponet}, is grounded in the universal approximation theorem for operators. Concurrently, FNO and its variants~\cite{li21-iclr-fno, wen22-awr-u-fno, guibas22-iclr-adaptivefourierneuraloperators, george24-tmlr-incremental-fno, pathak23-pasc-fourcastnet, tran23-iclr-factorized-fno} approximate integration with linear projection in the Fourier domain. GNO and its variants~\cite{li20-iclr-workshop-gno, li20-nips-mgno, li23-jmlr-neuraloperatorgraph} utilize graph neural networks (GNNs) as the base network, adapting neural operators to finite elements and complex mesh domains. The Transformer~\cite{vaswani17-nips-transformer} architecture, as a crucial milestone of deep learning, is also employed in the NO framework~\cite{xiao24-icml-ono, li23-nips-scalable-transformer, wu24-icml-transolver, mao24-icml-suggogate, liu23-corr-htnet}. To address the quadratic complexity of the attention mechanism, GNOT~\cite{hao23-icml-gnot} and ONO~\cite{xiao24-icml-ono} utilize the well-established linear Transformers~\cite{kitaev20-iclr-reformer, choromanski21-iclr-performer, cao21-nips-galerkin}. As another alternative, work like~\cite{cheng24-corr-mambaneuraloperatorwins, zheng24-nips-aliasfreemambaneuraloperator} attempted to employ the Mamba~\cite{gu23-colm-mamba, gu24-icml-mamba2} architecture as an alternative to Transformers in the NO framework. 

\paragraph{Neural PDE Solvers with Geometric Rigor} Many neural PDE solvers incorporate geometric rigor and deliver superior performance on regular grids. Greenfeld et al.~\cite{Greenfeld19} proposed to learn the prolongation operators of the geometric multigrid. UGrid~\cite{han24-icml-ugrid} proposed a learnable multigrid solver with a convergence guarantee, and adapts to arbitrary Dirichlet boundaries. For NOs, however, although the foundational work FNO~\cite{li21-iclr-fno} was proposed for regular grids, its successors focused on extensions to other domains, leaving regular grids largely unexplored. Most existing NO frameworks, both Transformer-based and Mamba-based, lack geometric rigor and mathematical insights, and fail to deliver optimal performance on regular grids.

\paragraph{State-Space Model (SSM) and Mamba} SSM~\cite{gu21-nips-lsslayer, gu21-iclr-efficiently, gupta22-nips-dssm, li22-corr-makes} is a mathematical framework representing time-variant systems by defining hidden states and their transitions. Mamba \cite{gu23-colm-mamba} introduced a selective mechanism and a hardware-aware implementation to SSMs, delivering a modeling capability in par with Transformers yet only with a linear complexity. When applied to the vision domain, Mamba-based methods employ a multiway cross-scan pattern~\cite{yang24-bmvc-plainmamba, xie24-nips-quadmamba, liu24-nips-vmamba}. However, this pattern lacks mathematical rigor, involves duplicate components in the hidden state, and decreases performance. Moreover, existing Mamba-based methods model one-dimensional sequences, requiring two-dimensional input to be flattened before being processed. This results in a loss of geometric information and also leads to sub-optimal performance. As a remedy, 2DMamba~\cite{zhang25-cvpr-2dmamba} manifests a two-dimensional state space representation, which performs well on the 2D domain for vision tasks. However, its potential in scientific computation remains unexplored. In summary, geometric rigor is an unexplored topic for Mamba-based NOs.

\section{Mathematical Preliminaries}

\paragraph{Neural Operators} The primary objective of a NO is to learn a mapping between two infinite-dimensional function spaces via a finite array of observed input-output pairs observed via numerical simulation, real-world applications, etc. This problem can be mathematically formulated as follows~\cite{li21-iclr-fno}: 

We suppose that $\Omega \subset \mathbb{R}^d$ is an open and bounded $d$-dimensional domain. We then let $\mathcal{A} = \mathcal{A}(\Omega; \mathbb{R}^{d_a})$ denote a Banach space of vector-valued functions mapping $\Omega \to \mathbb{R}^{d_a}$, and $\mathcal{U} = \mathcal{U}(\Omega; \mathbb{R}^{d_u})$ denote another Banach space of vector-valued  functions $\Omega \to \mathbb{R}^{d_u}$. We further let $\mathcal{U}^{*}$ be the dual space of $\mathcal{U}$. We then consider a generic family of PDEs in the form of: 
\begin{equation}
    \left\{
    \begin{aligned}
        (\mathcal{D}_a u) (x) & = f(x), & x \in \Omega \\
        u(x) & = 0, & x \in \partial \Omega
    \end{aligned}
    \right.
    \ .
    \label{equ:generic_family_of_pdes}
\end{equation}
Here, $a \in \mathcal{A}$ is a parameter; $\mathcal{D}_a: \mathcal{A} \to \mathcal{D}(\mathcal{U}; \mathcal{U}^{*})$ is a mapping from the parameter Banach space $\mathcal{A}$ to the space of linear differential operators $\mathcal{D}$ in the form of $\mathcal{U} \to \mathcal{U}^{*}$; and $f \in \mathcal{U}^{*}$ is the right-hand side. Eq.~\ref{equ:generic_family_of_pdes}'s solution $u: \Omega \to \mathbb{R}$ is a scalar-valued function living in the Banach space $\mathcal{U}$. We further let $G^{\dag} := {\mathcal{D}_a}^{-1} f: \mathcal{A} \to \mathcal{U}$ be an \textit{solution operator} mapping the parameter $a$ to the solution $u$: $a \mapsto u$. We assume that we have $N$ i.i.d. observations pairs $\{ (a_j, u_j) \}_{j=1}^{N}$, where $a_j$ is drawn from a probability measure $\mu$ supported on $\mathcal{A}$, and $u_j = G^{\dag}(a_j) + \varepsilon_j$ is potentially corrupted with noise $\varepsilon_j$. In the training process, we aim to construct an approximation of $G^{\dag}$ by constructing a parametric mapping $G: \mathcal{A} \times \mathbb{R}^p \to \mathcal{U}$, or equivalently, $G_{\theta}: \mathcal{A} \to \mathcal{U}, \ \theta \in \mathbb{R}^p$, by choosing the optimal parameter $\theta^{\dag} \in \mathbb{R}^p$ such that $G(\cdot, \theta^{\dag}) = G_{\theta^{\dag}} \approx G^{\dag}$. This formulation is equivalent to an optimization problem in infinite-dimensional spaces, with the cost functional $L: \mathcal{U} \times \mathcal{U} \to \mathbb{R}$ defined as $\min_{\theta \in \mathbb{R}^p} \mathbb{E}_{a \sim \mu} \left[ L(G_{\theta}(a),  G^{\dag}(a)) \right]$. 

\paragraph{State-Space Models and Mamba} An SSM manifests a time-evolving linear system that maps a one-dimensional input sequence $\vb{x}(t)$ to an output response sequence $\vb{y}(t)$ recurrently through $N$ mutually-independent hidden states $\vb{h}_{s}(t)$, $s = 1, 2, \cdots, N$. For clarity, in this paper, we refer to the $t$ dimension as ``\textbf{\textit{state}}'' dimension, and $s$ as ``\textbf{\textit{dstate}}'' dimension. Mamba~\cite{gu23-colm-mamba, gu24-icml-mamba2} introduces a selective mechanism to allow the SSMs to dynamically adapt to the input context. Mathematically, the continuous-form of the Mamba-SSM is formulated as Eq.~\ref{equ:2dmano-selective-sse}: \\
\quad
\begin{minipage}{0.47\textwidth}
\small
\begin{equation}
    \left\{
    \begin{aligned}
        \vb{h}_{s}'(t) & = \vb{A}_{s}(t) \vb{h}_{s}(t) + \vb{B}_{s}(\vb{x}(t)) \vb{x}(t) \ ,
        \\
        \vb{y}(t) & = \sum_{s = 1}^{N} \vb{C}_s(\vb{x}(t)) \vb{h}_{s}(t) \ .
    \end{aligned}
    \right.
    \label{equ:2dmano-selective-sse}
\end{equation}
\end{minipage}
\qquad
\begin{minipage}{0.42\textwidth}
\small
\begin{equation}
    \left\{
    \begin{aligned}
        \vb{h}_{s}[t] & = \overline{\vb{A}_{s}}[t] \vb{h}_{s}[t - 1] + \overline{\vb{B}_{s}}[t] \vb{x}[t] \ ,
        \\
        \vb{y}[t] & = \sum_{s = 1}^{N} \vb{C}_{s}[t] \vb{h}_{s}[t] \ .
    \end{aligned}
    \right.
    \label{equ:2dmano-time-invariant-sse-discretized}
\end{equation}
\end{minipage}

In Eq.~\ref{equ:2dmano-selective-sse}, $\vb{A}_s(t)$, $\vb{B}_s(\vb{x}(t))$, and $\vb{C}_s(\vb{x}(t))$ are learnable model parameters, with $\vb{B}_s$ and $\vb{C}_s$ being input-dependent. Following the zero-order hold (ZOH) rule, Eq.~\ref{equ:2dmano-selective-sse} could be discretized into Eq.~\ref{equ:2dmano-time-invariant-sse-discretized}. In Eq.~\ref{equ:2dmano-time-invariant-sse-discretized}, $\overline{\vb{A}_{s}}[t] = \exp(\Delta[t] \vb{A}_{s}(t))$, $\overline{\vb{B}_{s}}[t] = \Delta_t \vb{B}_{s}(\vb{x}(t))$, $\overline{\vb{C}_{s}}[t] = \vb{C}_{s}(\vb{x}(t))$, and $\Delta[t] = \mathrm{softplus}(\Delta(\vb{x}(t))) = \mathrm{ln}(1 + \exp(\Delta(\vb{x}(t))))$, where $[t]$ denotes the discrete value at time step $t$, $\mathrm{ln}(\cdot)$ denotes element-wise natural logarithm, and $\Delta(\vb{x}(t))$, $\vb{B}_{s}(\vb{x}(t))$, and $\vb{C}_{s}(\vb{x}(t))$ are linear functions of the input $\vb{x}(t)$. $\overline{\vb{A}_{s}}$, $\overline{\vb{B}_{s}}$, and $\overline{\vb{C}_{s}}$ jointly constitute the learnable parameters of the Mamba-SSM, whereas $\Delta[t]$ represents the time step granularity of the input-adaptive discretization process. 

The correctness of the NO framework itself, as well as that of employing Mamba in NOs, is a well-studied problem with mathematically-rigorous proofs~\cite{zheng24-nips-aliasfreemambaneuraloperator}. And thus, these are not our major concerns. For the theoretical completeness of this paper, we will discuss these mathematical details in Sec.~\ref{sec:appendix:mathematical-insights-mamba-nos}. 

\paragraph{2DMamba} In contrast to the vanilla Mamba, 2DMamba~\cite{zhang25-cvpr-2dmamba} manifests a two-dimensional hidden state, aggregating not only semantic, but also geometric information directly from a 2D domain. 2DMamba's coefficients remain the same as 1D, with the subscript being $[i, j]$ to index discretized 2D inputs instead of $[t]$, and is formulated as follows: 
\begin{equation}
    \small
    \left\{
    \begin{aligned}
        \vb{g}_s[i, j] & = \overline{\vb{A}_s}[i, j] \ \vb{g}[i, j - 1] + \overline{\vb{B}_s}[i, j] \ \vb{x}[i, j] 
        \ , \\
        \vb{h}_s[i, j] & = \overline{\vb{A}_s}[i, j] \ \vb{h}[i - 1, j] + \vb{g}_s[i, j]
        \ , \\
        \vb{y}[i, j] & = \sum_{s=1}^{N} \overline{\vb{C}_s}[i, j] \ \vb{h}_s[i, j]
        \ .
    \end{aligned}
    \right.
    \label{equ:2d-selective-sse-discretized}
\end{equation}
On a 2D domain, 2DMamba first aggregates a horizontal hidden state $\vb{g}$, which is then aggregated vertically into the final hidden state $\vb{h}$. 2DMamba is highly efficient due to its hardware-aware design. Yet, it is only applied to the vision domain, and its potential in scientific computation remains unexplored.

\section{Novel Method}

To improve neural operators on their accuracy, efficiency, and enhance their geometric rigor, we present our novel GeoMaNO framework. GeoMaNO manifests a geometric-aware architecture tailored for regular grids. As illustrated in Fig.~\ref{fig:geomano-arch}, GeoMaNO constructs the following iterative mapping from input $a$ to solution $u$:
\begin{align}
    a 
    \underbrace{\mapsto v_0}_{\text{encode}} 
    \underbrace{\mapsto v_1 \mapsto \cdots \mapsto v_T}_{T \text{ GeoMaNO layers}} 
    \underbrace{\mapsto u}_{\text{decode}}  
    \ ,
\end{align}
which is conducted by our geometry-aware encoder layer $\mathcal{E}$ (Sec.~\ref {sec:method:encoder}), our GeoMaNO layers $M_{1 \cdots T}$ (Sec.~\ref{sec:method:geomano-layer}), and our decoder layer $\mathcal{D}$ (Sec.~\ref{sec:method:encoder}). Depending on the dimensionality of the underlying physical domain~\footnote{E.g., the Navier-Stokes problem manifests a three-dimensional physical domain $(x, y, t)$, whereas the Darcy flow problem manifests a two-dimensional physical domain $(x, y)$.}~, our geometry-aware encoder and the GeoMaNO layers will behave differently, with the details to be discussed in the corresponding subsections. 

\subsection{Geometry-Aware Encoder and Decoder}
\label{sec:method:encoder}

\paragraph{Geometry-Aware Encoder Layer} The encoder layer seamlessly combines the legacy lifting operator~\cite{Kovachki23-jmlr-no} with the patch embedding operator~\cite{trockman23-iclr-patch-embedding}. The lifting operation is conducted by a feed-forward network~\cite{bebis94-ieee-feedforward}, which lifts a lower-dimensional input to a higher-dimensional latent representation. This effectively maps a potentially non-linear physical-domain input into a linear representation that resides in a higher-dimensional potential space~\cite{bevanda21-anc-koopman-operator}. Because we are focusing on regular grids, we further adopt the patch embedding~\cite{trockman23-iclr-patch-embedding} technique, which is extensively employed in the vision domain~\cite{jaegle21-icml-perceiver}. Mathematically, the geometry-aware encoding layer $\mathcal{E}$ could be formulated as:
\begin{equation}
    \vb{Y} = \mathrm{Linear}(\vb{X}), \quad
    \vb{W} = \mathrm{Softmax}(\mathrm{Linear}(\vb{Y})), \quad
    \vb{Z} = \mathrm{LayerNorm}(\vb{W}^{\top} \vb{Y}),
\end{equation}
where $\mathrm{Linear}(\cdot)$ denotes a standard biased feed-forward network with activation, $\mathrm{Softmax}(\cdot)$ denotes feature-wise softmax, and $\mathrm{LayerNorm}(\cdot)$ denotes the layer normalization~\cite{ba16-iclr-layernorm}. Here, $\vb{X} \in \mathbb{R}^{P \times D_{i}}$ denotes the physical-domain input with $P$ grid points, and $D_{i}$ is the depth of the input~\footnote{E.g., for the Darcy flow problem $\nabla \cdot (f(x, y) \nabla u(x, y)) = g(x, y)$, $\vb{X}$ encodes the coeffieients $f$ and the right-hand side $g$. For this example, we have $D_i = 2$.}~. $\vb{X}$ is first lifted to $\vb{Y} \in \mathbb{R}^{P \times D}$, where $D \gg D_{i}$, then embedded into $L$ latent patches, with $P \ll L$. Inspired by Mamba's~\cite{gu23-colm-mamba, gu24-icml-mamba2} input-adaptive design, our encoding process is conducted with an input-adaptive set of parameters $\vb{W} \in \mathbb{R}^{P \times L}$, $\vb{W} = \vb{W}(\vb{X})$. 

For PDEs on 1D or 3D physical domains, we adopt the regular patch embedding pipeline, which embeds positional encodings into the latent patches. However, for PDEs on 2D domains, we do \textbf{\textit{not}} embed the positional encodings. In both cases, the input is \textbf{\textit{not}} flattened, and the geometric structure is maintained for postprocessing. This geometry-aware patch embedding operation preserves the structured geometry and the spatial relationship between latent patches for the succeeding GeoMaNO layers. 

\paragraph{Decoder Layer} The decoder layer $\mathcal{D}$ can be viewed as the inverse of the encoding layer $\mathcal{E}$. It converts the latent patches back to the physical domain. Mathematically, $\mathcal{D}$ could be formulated as: 
\begin{equation}
    \vb{W} = \mathrm{Softmax}(\mathrm{Linear}(\vb{Z})), \quad 
    \vb{Y} = \vb{W}^{\top} \vb{Z}, \quad 
    \vb{X} = \mathrm{LayerNorm}(\mathrm{Linear}(\vb{Y})), 
\end{equation}
where the latent patches $\vb{Z} \in \mathbb{R}^{L \times D}$ are projected back to the physical grid $\vb{X} \in \mathbb{R}^{P}$. 

\subsection{GeoMaNO Layer}
\label{sec:method:geomano-layer}

As illustrated in Fig.~\ref{fig:geomano-arch} (middle), our GeoMaNO layer parallels the legacy self-attention~\cite{vaswani17-nips-transformer} block, but we use pre-norms instead of post-norms to preserve the skip-connection gradients: 
\begin{equation}
    \begin{aligned}
        \vb{S}_{t} &= \vb{Z}_{t-1} + \mathrm{GeoMambaKernel}(\mathrm{LayerNorm}(\vb{Z}_{t-1})), \\
        \vb{Z}_{t} &= \vb{S}_{t} + \mathrm{Linear}(\mathrm{LayerNorm}(\vb{S}_{t})). 
    \end{aligned}
\end{equation}

\begin{figure}[ht]
    \centering
    \includegraphics[width=\linewidth]{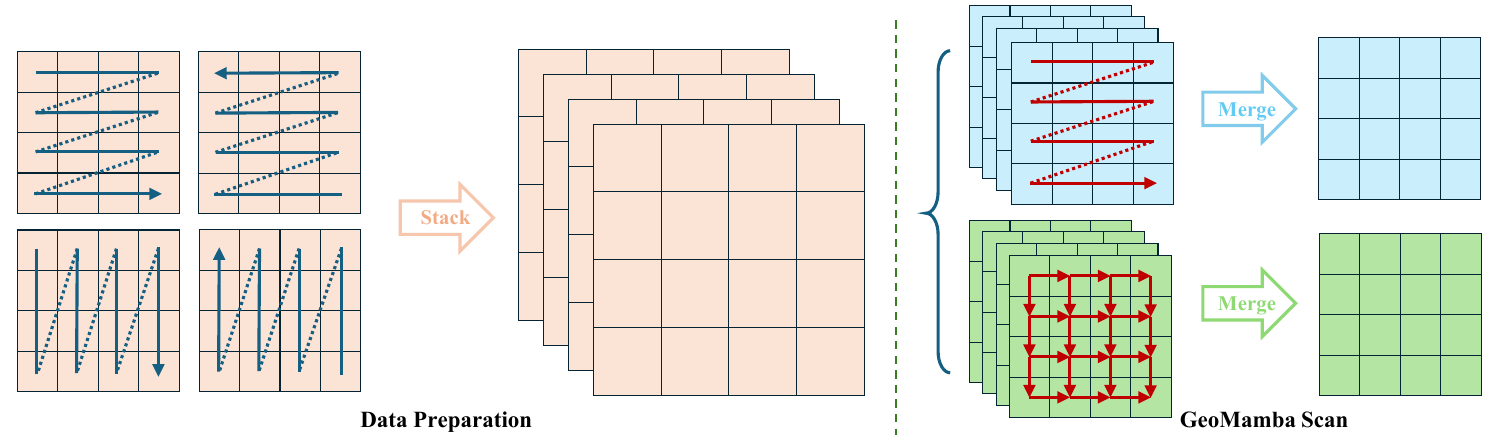}
    \vspace{-10pt}
    \caption[Four-way cross-scan.]{The four-way cross-scan pattern. \textbf{Left}: In the data preparation stage, the input grid is traversed in four orders, and the traversal results are stacked at an additional order dimension. \textbf{Right}: For each slice along the order dimension, we process it with our GeoMamba-SSM module (shown in {\color{blue} blue} color). As a special case, for 2D PDEs, in GeoMamba-SSM, we employ a 2D SSM representation with geometric rigor (shown in {\color{green} green} color). The results for all four cross-scan directions are merged.}
    \label{fig:cross-scan-directions}
\end{figure}

\paragraph{GeoMamba Kernel} Following the common practice employed by the vision domain~\cite{liu24-nips-vmamba, zhu24-icml-vim}, we adopt a \ \textbf{\textit{four-way cross-scan pattern}} over the input, which has been proven effective on regular grids. Given a 2D input grid (for 3D input domains, we treat the time dimension as an additional feature dimension), we traverse the input in four orders: (1) top-left to bottom-right (row-major), (2) bottom-right to top-left (row-major), (3) top-left to bottom-right (column-major), and (4) bottom-right to top-left (column-major). The traversal results are stacked at a new dimension, as illustrated in Fig.~\ref{fig:cross-scan-directions} (left half).

The GeoMamba-SSM regresses each scan direction independently, and the results for the four scan directions are merged. Our novel GeoMamba kernel could be formulated as Algorithm~\ref{alg:geomamba-kernel}, where $B$ denotes the batch size, $L$ denotes the size of the input patch grid, $ED$ denotes the feature embedding dimension, $N$ denotes the Mamba state dimension, $\sigma$ denotes element-wise activation, and $\circ$ denotes element-wise multiplication. In Algorithm~\ref{alg:geomamba-kernel}, lines 3-6 correspond to the left half of Fig.~\ref{fig:cross-scan-directions}, whereas lines 8-9 correspond to the right half of Fig.~\ref{fig:cross-scan-directions}.  

\begin{algorithm}[ht]
    \caption{GeoMamba kernel}
    \label{alg:geomamba-kernel}
    \begin{algorithmic}[1]
        \small
        \Require{Input patch grid $\vb{Z}: (B, L, ED)$}; 
        \Require{GeoMamba-SSM scanning coefficients $\vb{As}: (N, ED)$, $\vb{Rs}: (N, ED)$};
        \Ensure{Output patch grid $\vb{Y}: (B, L, ED)$}.
        
        \State $\vb{Z} = \mathrm{Linear}(\vb{Z})$; \Comment{Input projection.}
        \State $\vb{X} = \sigma(\mathrm{Conv}(\vb{Z}))$;

        \For {Scan directions $d = 1, 2, 3, 4$} \Comment{Cross-scan data preparation.}
            \State $\vb{X}_d = \mathrm{Traverse}_d(\vb{X})$;
        \EndFor

        \State $\vb{Xs} = (\vb{X}_1, \vb{X}_2, \vb{X}_3, \vb{X}_4)$;
        \State $\vb{\Delta s}, \vb{Bs}, \vb{Cs} = \mathrm{Linear}(\vb{Xs})$;

        \State $\vb{Y}_1, \vb{Y}_2, \vb{Y}_3, \vb{Y}_4 = \mathrm{GeoMamba}\text{-}\mathrm{SSM}(\vb{Xs}, \vb{\Delta s}, \vb{As}, \vb{Bs}, \vb{Cs}, \vb{Rs})$; \Comment{Parallel cross-scan.} 
        \State $\vb{Y} = \mathrm{CrossMerge}(\vb{Y}_1, \vb{Y}_2, \vb{Y}_3, \vb{Y}_4)$;

        \State $\vb{Y} = \sigma(\mathrm{LayerNorm}(\vb{Y})) \circ \sigma(\vb{Z})$; 
        \State $\vb{Y} = \mathrm{Linear}(\vb{Y})$. \Comment{Output projection.}
\end{algorithmic}
\end{algorithm}

\subsection{GeoMamba-SSM Module} 
\label{sec:method:geomamba-ssm-module}

\paragraph{Theoretical Insights} Although proven effective on structured domains like regular grids, the four-way cross-scan pattern lacks geometric rigor: When the results from different scan orders are merged, it brings \textbf{\textit{duplicate latent information}} into the hidden states. Consider a simplified form of Eq.~\ref{equ:2dmano-time-invariant-sse-discretized}: $h_t = A h_{t-1} + B x_t$ (with $A$ and $B$ be constants for simplicity). When this recursion is applied to latent tokens on a regular grid (left), the hidden states (along the four scan directions) for $x_4$ are listed as follows (right): \\
\begin{minipage}{0.35\textwidth}
    \[
        \begin{pmatrix}
            x_0 & x_1 & x_2 \\
            x_3 & x_4 & x_5 \\
            x_6 & x_7 & x_8
        \end{pmatrix}
    \]
\end{minipage}
\quad
\begin{minipage}{0.42\textwidth}
    \small
    \[
    \begin{aligned}
        h_4^{(1)} &= A^4 B x_0 + A^3 B x_1 + A^2 B x_2 + AB x_3 + {\color{red} B x_4}, \\
        h_4^{(2)} &= A^4 B x_8 + A^3 B x_7 + A^2 B x_6 + AB x_5 + {\color{red} B x_4}, \\
        h_4^{(3)} &= A^4 B x_0 + A^3 B x_3 + A^2 B x_6 + AB x_1 + {\color{red} B x_4}, \\
        h_4^{(4)} &= A^4 B x_8 + A^3 B x_5 + A^2 B x_2 + AB x_7 + {\color{red} B x_4}, \\
    \end{aligned}
    \]
\end{minipage} 

where the superscripts $^{(1) \sim (4)}$ separately denote the four cross-scan directions. When the four scan directions are merged, the hidden state ${\color{red} B x_4}$ is duplicated four times. In practice, although $A$ and $B$ vary for scan directions and are input-dependent, the SSM still regresses over duplicate information. These duplicates pose an excessive optimization burden to the training process, and as shown in Table~\ref{tab:ablations}, will degrade the performance. 

\paragraph{GeoMamba-SSM} As a remedy to this challenge, we propose our novel GeoMamba-SSM formulation, which could be formulated as:
\begin{equation}
    \small
    \left\{
    \begin{aligned}
        \vb{h}_{s}^{(d)}[t] & = \overline{\vb{A}_{s}}^{(d)}[t] \ \vb{h}_{s}^{(d)}[t - 1] + \overline{\vb{B}_{s}}^{(d)}[t] \ \vb{x}^{(d)}[t] \ ,
        \\
        \vb{y}^{(d)}[t] & = \sum_{s = 1}^{N} \qty( \vb{C}_{s}^{(d)}[t] \ \vb{h}_{s}^{(d)}[t] {\color{blue} \ - \ \overline{\vb{Rs}}^{(d)}[t] \  \overline{\vb{B}_{s}}^{(d)}[t] \ \vb{x}^{(d)}[t]} ) \ , 
        \\ 
        \vb{y}[t] &= \mathrm{CrossMerge}(\vb{y}^{(d)}[t]) \ .
        \\
    \end{aligned}
    \right.
    \label{equ:1d-geomano-ssm}
\end{equation}
To dampen the duplicate hidden states, GeoMamba-SSM introduces a \textbf{\textit{geometric correction component}}, which is highlighted in {\color{blue} blue} color. The correction component manifests a configurable coefficient $\overline{\vb{Rs}}$, which could be either fixed or learnable. With coefficients fixed, the correction component is enforced as a hard constraint; otherwise, it will be a soft constraint. 

\begin{remark}[\textbf{Mamba's D Coefficients}]
    Some Mamba-SSM variants allow an extra coefficient $\vb{D} = (D_1, \cdots, D_N) \in \mathbb{R}^{N}$, which will rewrite the second line of Eq.~\ref{equ:2dmano-time-invariant-sse-discretized} into $\vb{y}[t] = \sum_{s = 1}^{N} \qty( {\vb{C}_{s}}[t] {\vb{h}_{s}}[t] {\color{blue} \ + \  D_s \  \vb{x}[t]} )$. However, per \textbf{\textit{dstate}} dimension $s$, $D_s$ is a scalar constant across the entire grid, and this residual offers no input-adaptive control over $B$. In contrast, GeoMamba-SSM introduces input-adaptive residuals explicitly on $Bx$: see Eq.~\ref{equ:1d-geomano-ssm} for the difference. The $\vb{D}$ coefficients are \textbf{\textit{already integrated}} into our GeoMamba-SSMs. As shown in Table~\ref{tab:ablations}, compared with $\vb{D}$-only versions, GeoMaNO performs better when it also has $\overline{\vb{Rs}}$ enabled. 
\end{remark} 

\vspace{5pt}
\begin{remark}[\textbf{Design Choices for Fixed Geometric Correction Coefficients}]
    \label{thm:geo-correction-coeffs}
    We employ a scan-order-wise constant implementation for fixed correction coefficients. For each scan direction, we set $\overline{\vb{Rs}}$ to either $0$ or $1$ for the entire grid. Since $Bx$ is duplicated four times, it would be beneficial to dampen one, two, or three copies of $Bx$. We ablate $\overline{\vb{Rs}}$ with three configurations: $0001$, $0011$, or $0111$, where the four digits correspond to the four scan directions. Results are available in Table~\ref{tab:ablations}. 
\end{remark}

\begin{figure*}[ht]
    \centering
    \includegraphics[width=0.76\linewidth]{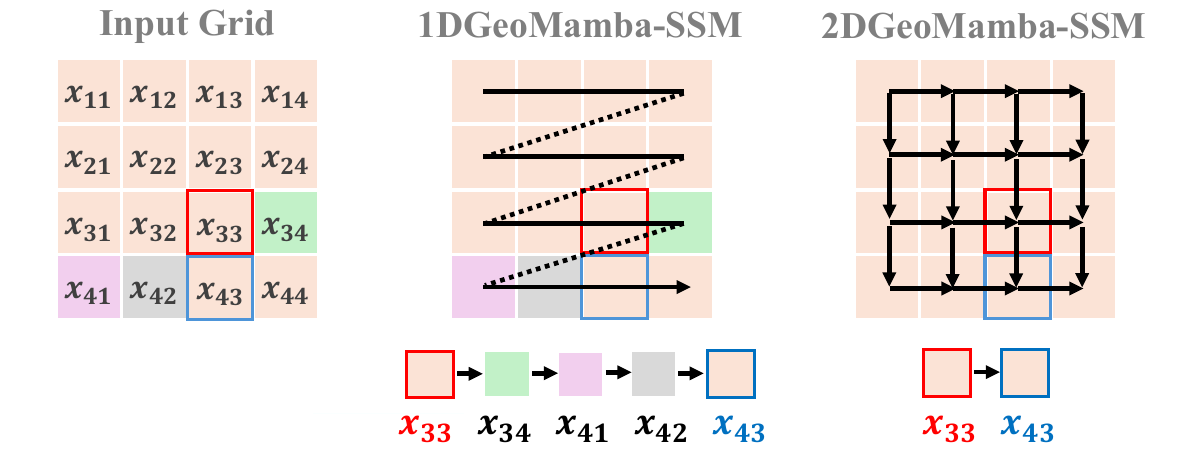}
    \vspace{-2pt}
    \caption[GeoMamba-SSM variants.]{Illustration of our GeoMamba-SSM variants. \textbf{Left}: A sample $4 \times 4$ input grid. \textbf{Middle}: 1DGeoMamba-SSM flattens the input into a 1D sequence and regresses the hidden states over it. However, after flattening, vertically-adjacent patches {\color{red} $x_{33}$} (highlighted with {\color{red} red} margins) and {\color{blue} $x_{43}$} (highlighted with {\color{blue} blue} margins) are no longer adjacent. This leads to \textbf{\textit{geometric inconsistency}} in the hidden states. \textbf{Right}: 2DGeoMamba-SSM regresses over the input grid in a 2D manner. Adjacent patches remain adjacent in the hidden states, properly preserving geometric information.}
    \label{fig:main-spatial-discrepancy}
\end{figure*}

\paragraph{2DGeoMamba-SSM} The vanilla Mamba-SSM employs a general-purpose one-dimensional state-space representation, which, in principle, generalizes to arbitrary dimensions and domains. However, being generic indicates the vanilla Mamba-SSM may not perform optimally under domain-specific settings, e.g., 2D problems on regular grids. As illustrated in Fig.~\ref{fig:main-spatial-discrepancy} (middle), 1DGeoMamba-SSM leads to \textbf{\textit{geometric inconsistency}} in the hidden states. To further enhance GeoMamba-SSM's geometric rigor, for 2D problems, we specialize it with a two-dimensional state-space representation, as inspired by~\cite{zhang25-cvpr-2dmamba}. Our 2DGeoMamba-SSM is formulated as:
\begin{equation}
    \small
    \left\{
    \begin{aligned}
        \vb{g}_s^{(d)}[i, j] & = \overline{\vb{A}_s}^{(d)}[i, j] \ \vb{g}_s^{(d)}[i, j - 1] + \overline{\vb{B}_s}^{(d)}[i, j] \ \vb{x}^{(d)}[i, j] 
        \ , \\
        \vb{h}_s^{(d)}[i, j] & = \overline{\vb{A}_s}^{(d)}[i, j] \ \vb{h}_s^{(d)}[i - 1, j] + \vb{g}^{(d)}_s[i, j] \ , 
        \\ 
        \vb{y}^{(d)}[i, j] & = \sum_{s=1}^{N} \qty( \overline{\vb{C}_s}^{(d)}[i, j] \ \vb{h}_s^{(d)}[i, j] {\color{blue} \ - \  \overline{\vb{Rs}}^{(d)}[i, j] \ \overline{\vb{B}_{s}}^{(d)}[i, j] \ \vb{x}^{(d)}[i, j]} )
        \ , \\ 
        \vb{y}[i, j] &= \mathrm{CrossMerge}(\vb{y}^{(d)}[i, j]) 
        \ , \\
    \end{aligned}
    \right.
    \label{equ:2d-geomano-ssm}
\end{equation}
where the notations remain the same as Eq.~\ref{equ:1d-geomano-ssm} and Eq.~\ref{equ:2d-selective-sse-discretized}. 2DGeoMamba-SSM conducts a \textbf{\textit{two-dimensional cross-scan over the input grid}}, as illustrated in Fig.~\ref{fig:main-spatial-discrepancy} (right). 2DGeoMamba-SSM preserves the neighboring relations in the hidden states, and as shown in Table~\ref{tab:ablations} (a), improves the performance. More mathematical insights will be discussed in Sec.~\ref{sec:appendix:theoretical-insights-2dmamba}.

\section{Experiments}
\label{sec:experiments}

\subsection{System Setup}

\paragraph{Benchmarks and Baselines} We evaluate GeoMaNO's performance on regular grids with the standard neural operator PDE solving benchmarks~\cite{thuml25-online-pde-solving-std-benchmark}. We use the Darcy flow benchmark, which is a two-dimensional linear elliptic problem, and the Navier-Stokes benchmark, which is a three-dimensional (including the time dimension) non-linear parabolic/hyperbolic problem. We compare GeoMaNO with 14 baselines, with more details available in Sec.~\ref{sec:appendix:system-setup}. 

\paragraph{Implementation Details} We follow the standard configuration employed by Transolver~\citep{wu24-icml-transolver}. For the main results, we set the model depth $T$ to 8 layers, and the feature embedding dimension $ED$ to 64 for Darcy flow and 256 for Navier-Stokes. Regarding the GeoMaNO layers, we employ the vanilla GeoMamba-SSM for Navier-Stokes and 2DGeoMamba-SSM for Darcy flow. Our experiments are conducted on one NVIDIA Quadro RTX 8000 GPU, and repeated five times for consistency. More details will be discussed in Sec.~\ref{sec:appendix:system-setup}. 

\subsection{Main Results}

\paragraph{Standard Benchmarks} We report GeoMaNO's performance on standard benchmarks~\cite{thuml25-online-pde-solving-std-benchmark}. As shown in Table~\ref{tab:mainres_standard}, GeoMaNO achieves consistent SOTA performance across both benchmarks, covering both linear elliptic problems and non-linear parabolic/hyperbolic problems. GeoMaNO surpasses the current SOTA, Transolver~\cite{wu24-icml-transolver}, by as much as $58.9\%$ on the Navier-Stokes benchmark and $36.8\%$ on the Darcy flow benchmark. These results highlight GeoMaNO's superior performance in PDE solution operator approximation. 

\begin{table*}[ht]
    \small
    \caption{Performance comparison on standard benchmarks. We report the relative L2 error (the smaller, the better). The best result is highlighted in \textbf{bold}, whereas the second best is highlighted \underline{\textit{underlined italic}}. For clarity, we also report our relative performance gain $(1 - \frac{\text{our error}}{\text{2nd-best error}})\%$.}
    \label{tab:mainres_standard}
    \centering

    \renewcommand{\multirowsetup}{\centering}
    \setlength{\tabcolsep}{7.2pt}
    
    \begin{subtable}{0.48\textwidth}
        \centering
        \begin{tabular}{l|cc}
            \toprule
                \multirow{3}{*}{Model} & \multicolumn{2}{c}{Relative L2 ($\downarrow$)} \\
                \cmidrule(lr){2-3} & \ N.-S.\quad & Darcy Flow \\
            \midrule
                FNO~\citep{li21-iclr-fno} & 0.1556 & 0.0108 \\
                WMT~\citep{gupta21-nips-multiwavelet} & {0.1541} & 0.0082 \\
                U-FNO~\citep{wen22-awr-u-fno} & 0.2231 & 0.0183 \\
                Geo-FNO~\citep{li23-jmlr-fno-general-geometry} & 0.1556 & 0.0108 \\
                U-NO~\citep{rahman23-tmlr-u-no} & 0.1713 & 0.0113 \\
                F-FNO~\citep{tran23-iclr-factorized-fno} & 0.2322 & {0.0077} \\
                LSM~\citep{wu23-icml-lsm} & 0.1535 & 0.0065 \\
            \midrule
                - \phantom{\textbf{GeoMaNO(Ours)}} & - & - \\
                - & - & - \\
            \bottomrule
        \end{tabular}
    \end{subtable}
    \hspace{0.02\textwidth}
    \begin{subtable}{0.48\textwidth}
        \centering
        \begin{tabular}{l|cc}
            \toprule
                \multirow{3}{*}{Model} & \multicolumn{2}{c}{Relative L2 ($\downarrow$)} \\
                \cmidrule(lr){2-3} & \ N.-S. \quad & Darcy Flow \\
            \midrule
                Galerkin~\citep{cao21-nips-galerkin} & 0.1401 & 0.0084 \\
                HT-Net~\citep{liu23-corr-htnet} & 0.1847 & 0.0079 \\
                OFormer~\citep{li23-tmlr-transformer-pde} & 0.1705 & 0.0124 \\
                GNOT~\citep{hao23-icml-gnot} & 0.1380 & 0.0105 \\
                FactFormer~\citep{li23-nips-fastformer} & 0.1214 & 0.0109 \\
                ONO~\citep{xiao24-icml-ono} & 0.1195 & 0.0076 \\
                Transolver~\citep{wu24-icml-transolver} & \underline{\textit{0.0900}} & \underline{\textit{0.0057}} \\
            \midrule
                \textbf{GeoMaNO (Ours)} & \textbf{0.0370} & \textbf{0.0036} \\
                Performance Gain & 58.9\% & 36.8\% \\
            \bottomrule
        \end{tabular}
    \end{subtable}
\end{table*}

\paragraph{Efficiency} We evaluate GeoMaNO's efficiency against the current SOTA, Transolver~\cite{wu24-icml-transolver}. In our evaluation, we employ model configurations that achieve the results in Table~\ref{tab:mainres_standard}. We report the training time $T_{\mathrm{train}}$ (seconds per epoch), inference time $T_{\mathrm{infer}}$ (seconds per epoch), and the model's GPU memory consumption (MB). For all metrics, the smaller, the better. The epoch size is kept consistent between both models. The results are available in Table~\ref{tab:efficiency}. GeoMaNO consistently outperforms Transolver~\cite{wu24-icml-transolver} over all metrics and on both benchmarks, notably by $49.8\%$ for Darcy flow inference. The results highlight GeoMaNO's superior computational efficiency. 

\begin{table*}[ht]
    \small
    \caption{Efficiency comparison of GeoMaNO and the current SOTA, Transolver~\cite{wu24-icml-transolver}.}
    \label{tab:efficiency}
    \vspace{-5pt}
    \centering

    \begin{subtable}{0.48\textwidth}
        \centering
        \subcaption{Navier-Stokes}
        \vspace{-3pt}
        
        \begin{tabular}{c|c|c|c}
            \toprule
                Model & $T_{\mathrm{train}}$ ($\downarrow$)  & $T_{\mathrm{infer}}$ ($\downarrow$)  & Mem. ($\downarrow$)  \\
            \midrule
                Transolver & 305.33 & 23.47 & 91.85 \\
            \midrule
                \textbf{GeoMaNO} & \textbf{268.68} & \textbf{16.61} & \textbf{70.44} \\
                Perf. Gain & 12\% & 29.2\% & 23.3\% \\
            \bottomrule
        \end{tabular}
    \end{subtable}
    \hspace{0.02\textwidth}
    \begin{subtable}{0.48\textwidth}
        \centering
        \subcaption{Darcy Flow}
        \vspace{-3pt}
        
        \begin{tabular}{c|c|c|c}
            \toprule
                Model & $T_{\mathrm{train}}$ ($\downarrow$)  & $T_{\mathrm{infer}}$ ($\downarrow$)  & Mem. ($\downarrow$) \\
            \midrule
                Transolver & 30.99 & 2.33 & 42.55 \\
            \midrule
                \textbf{GeoMaNO} & \textbf{23.98} & \textbf{1.17} & \textbf{5.21} \\
                Perf. Gain & 22.6\% & 49.8\% & 87.8\% \\
            \bottomrule
        \end{tabular}
    \end{subtable}
\end{table*}





\begin{figure}[ht]
    \centering
    \begin{subfigure}[h]{0.3\textwidth}
        \centering
        \includegraphics[width=\linewidth]{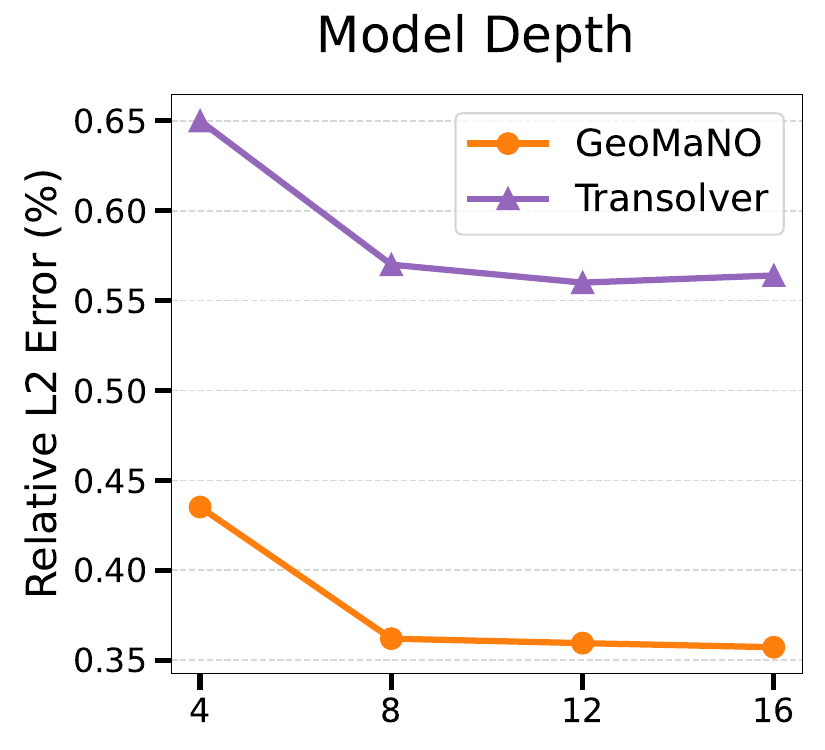}
    \end{subfigure}
    \begin{subfigure}[h]{0.3\textwidth}
        \centering
        \includegraphics[width=\linewidth]{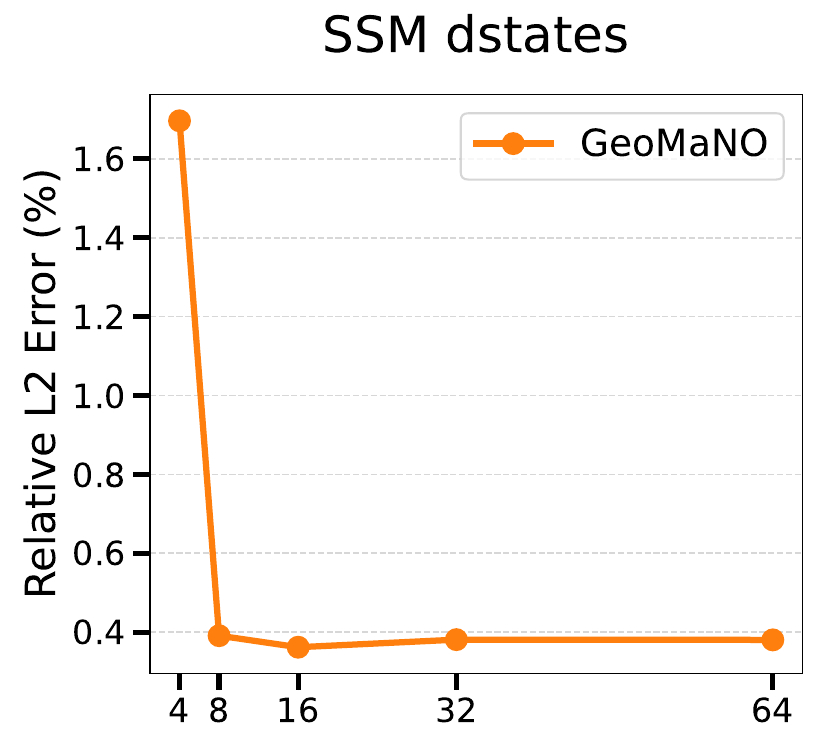}
    \end{subfigure}
    \begin{subfigure}[h]{0.3\textwidth}
        \centering
        \includegraphics[width=\linewidth]{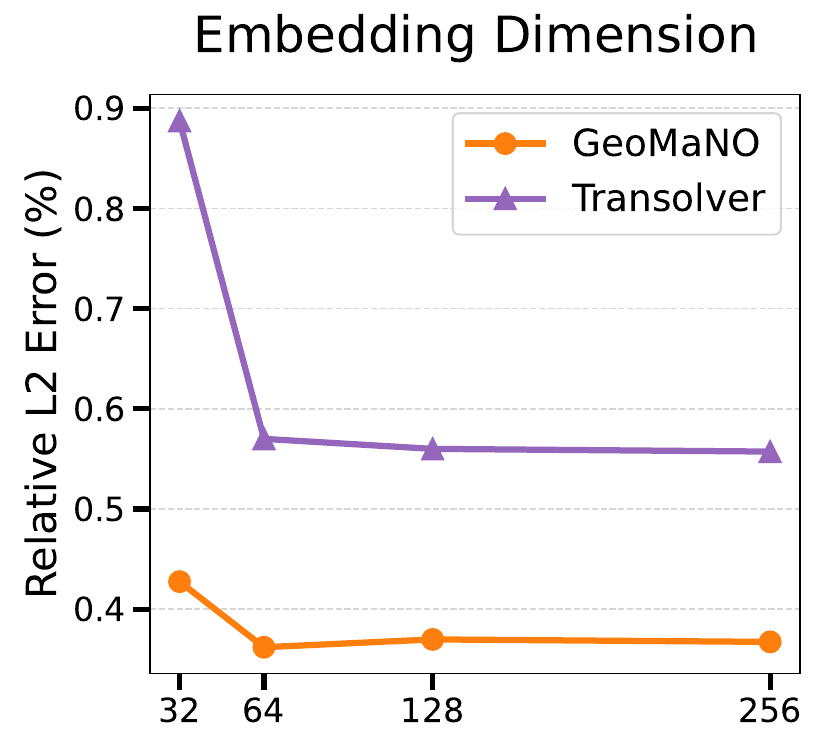}
    \end{subfigure}
    \caption[GeoMaNO's Scalability.]{GeoMaNO's scalability. We ablate model depth and embedding dimensions, and report the relative L2 errors ($\%$).}
    \label{fig:scalability}
\end{figure}

\paragraph{Scalability} To evaluate GeoMaNO's scalability, we ablate its model depth, the number of Mamba dstates, and the number of embedding dimensions on the Darcy flow benchmark. The results are illustrated in Fig.~\ref{fig:scalability}. It can be observed that GeoMaNO scales well to these three metrics when the hyperparameters are small. However, GeoMaNO won't scale infinitely: When these parameters go too large, the performance gain becomes less significant. 

\subsection{Ablation Studies}

We ablate GeoMaNO by the following factors: (1) Vanilla GeoMamba-SSM vs. 2DGeoMamba-SSM; (2) Positional encoding (none vs. yes); (3) The geometric correction coefficient $\vb{Rs}$ (none vs. fixed vs. learnable), and the quantitative results are available in Table~\ref{tab:ablations}. 

\paragraph{Darcy Flow} The Darcy flow problem models fluid pressure over a 2D regular grid. Its 2D physical nature calls for a NO specialized for the 2D domain. As shown in Table~\ref{tab:ablations} (a), 2DGeoMamba-SSM consistently outperforms the general-purpose vanilla GeoMamba-SSM. Among the 2D variants, the positional embedding decreases the performance. This is because positional embedding injects extra content into the grid and pollutes geometric information. The best performance is achieved with geometric corrections enabled, showcasing the necessity of our geometrically rigorous design. 

\paragraph{Navier-Stokes} The Navier-Stokes problem models fluid vorticity for a continuous period over a 2D regular grid. This problem is governed by a 3D physical domain $(x, y, t)$. The vanilla GeoMamba-SSM performs well, whereas 2DGeoMamba-SSM didn't converge well because it's specialized for 2D domains, and fails to capture the temporal dependencies. Again, the best performance is achieved with geometric correction enabled, highlighting the necessity of geometric rigor.  

\begin{table*}[ht]
    \small
    \caption{Ablation studies on the two benchmarks. We report the relative L2 error, and the rows for the best configurations are highlighted in \textbf{bold}. For the ``Geo. Coeff.'' columns, ``None'' denotes no geometric correction, ``Learnable'' denotes learnable geometric correction coefficients, and ``0001'' $\sim$ ``0111'' denotes fixed coefficients as detailed in \textit{Remark}~\ref{thm:geo-correction-coeffs}.}
    \label{tab:ablations}
    \vspace{-5pt}
    \centering

    \begin{subtable}{0.48\textwidth}
        \centering
        \subcaption{Darcy Flow}
        \vspace{-3pt}
        
        \begin{tabular}{c|c|c|c}
            \toprule
                SSM & Pos. Emb. & Geo. Coeff. & Rel. L2 ($\downarrow$) \\
            \midrule
                Vanilla & \checkmark & None & 0.00389 \\
            \midrule
                Vanilla & None & None & 0.00381 \\
                Vanilla & None & 0001 & 0.00385 \\
                Vanilla & None & 0011 & 0.00376 \\
                Vanilla & None & 0111 & 0.00387 \\
                Vanilla & None & Learnable & 0.00386 \\
            \midrule
                2D & \checkmark & None & 0.00428 \\
            \midrule
                2D & None & None & 0.00380 \\
                2D & None & 0001 & 0.00387 \\
                \textbf{2D} & \textbf{None} & \textbf{0011} & \textbf{0.00362} \\
                2D & None & 0111 & 0.00383 \\
                2D & None & Learnable & 0.00372 \\
            \bottomrule
        \end{tabular}
    \end{subtable}
    \hspace{0.02\textwidth}
    \begin{subtable}{0.48\textwidth}
        \centering
        \subcaption{Navier-Stokes}
        \vspace{-3pt}
        
        \begin{tabular}{c|c|c|c}
            \toprule
                SSM & Pos. Emb. & Geo. Coeff. & Rel. L2 ($\downarrow$) \\
            \midrule
                Vanilla & None & None & 0.04190 \\
                Vanilla & None & 0001 & 0.04212 \\
                Vanilla & None & 0011 & 0.04062 \\
                Vanilla & None & 0111 & 0.04153 \\
                Vanilla & None & Learnable & 0.04190 \\
            \midrule
                Vanilla & \checkmark & None & 0.03938 \\
                Vanilla & \checkmark & 0001 & 0.03795 \\ 
                Vanilla & \checkmark & 0011 & 0.03783 \\
                Vanilla & \checkmark & 0111 & 0.03709 \\ 
                \textbf{Vanilla} & \checkmark & \textbf{Learnable} & \textbf{0.03701} \\
            \midrule
                2D & None & None & 0.13554 \\
            \midrule
                2D & \checkmark & None & 0.18287 \\
            \bottomrule
        \end{tabular}
    \end{subtable}
\end{table*}

\paragraph{Generalization to Other Domains} We further evaluate GeoMamba-SSM's generalization to other domains. We integrate GeoMamba-SSM into the SOTA Mamba-based foundation framework, VMamba~\cite{liu24-nips-vmamba}, and evaluate it with the ImageNet-100 classification benchmark~\cite{deng09-cvpr-imagenet}. GeoMamba-SSM outperforms vanilla Mamba-SSM by $0.22\%$ on the top-1 accuracy score, which is significant in the natural image domain. More details are available in Sec.~\ref{sec:appendix:generalization-to-other-domains}. 

\section{Conclusion}
\label{sec:conclusion}

In this paper, we propose \textbf{GeoMaNO}, a Mamba-based neural operator framework with geometric rigor and mathematical insights. GeoMaNO manifests a geometric-aware architecture specifically tailored for regular grids, combined with our GeoMamba-SSM module, which employs a geometric correction component to dampen duplicate hidden states. GeoMaNO outperforms the SOTA in PDE solution operator approximation in both accuracy and efficiency. Extensive experiments validate all our claimed technical merits. 

\paragraph{Limitation and Future Research Directions} GeoMaNO currently works only on regular grids. However, in theory, our GeoMamba-SSM should also generalize to other unstructured domains like meshes and point clouds. Furthermore, the 2D version of our GeoMamba-SSM works well on 2D physical domains (Darcy benchmark) but fails to capture temporal dependencies on 3D domains (Navier-Stokes benchmark). It would be more beneficial to extend 2DMamba-SSM to 3D space. We would like to focus our future research efforts on these extensions and enhancements.


\newpage
\bibliographystyle{ieeetr}
\bibliography{bibliography}


\newpage
\appendix

\begin{center}
    {
    \Large
    \bf
    Supplementary Materials \\
    }
    \vspace{5pt}
    \large
    GeoMaNO: Geometric Mamba Neural Operator \\ 
    for Partial Differential Equations
\end{center}
\vspace{10pt}

\section{Specifications for Experiments}
\label{sec:appendix:system-setup}

\subsection{Benchmarks}

We evaluate GeoMaNO's performance on regular grids with the standard neural operator PDE solving benchmarks~\cite{thuml25-online-pde-solving-std-benchmark}, which is employed by most neural operator frameworks from FNO~\cite{li21-iclr-fno} to Transolver~\cite{wu24-icml-transolver}. We use the Darcy flow benchmark, which is a two-dimensional linear elliptic problem, and the Navier-Stokes benchmark, which is a three-dimensional (including the time dimension) non-linear parabolic/hyperbolic problem. 

\paragraph{Darcy Flow~\cite{li21-iclr-fno}} The Darcy Flow equation represents the steady-state diffusion process over porous media. The 2D Darcy Flow equation over a unit square is formulated as: 

\begin{equation}
    \left\{
    \begin{aligned}
        \nabla \cdot (a(x, y) \nabla u(x, y)) &= f(x, y), & (x, y) \in (0, 1)^2, \\ 
        u(x, y) &= 0, & (x, y) \in \partial (0, 1)^2, \\
    \end{aligned}
    \right.
    \label{equ:2dmano:darcy-flow}
\end{equation}

where $a(x, y)$ is the viscosity, $f(x, y)$ is the forcing term, and $u(x, y)$ is the solution. 
Further, Eq.~\ref{equ:2dmano:darcy-flow} is modified in the form of a temporal evolution as follows:


In this benchmark, the input is represented by the parameter $a(x, y)$, and the corresponding output is the solution $u(x, y)$. For consistency with FNO~\cite{li21-iclr-fno}'s evaluation, the coefficients $a(x, y)$ are generated according to a 2D Gaussian distribution: $a \sim \mathcal{N}(0, (-\Delta + 9\vb{I})^{-2})$, with zero Neumann boundary conditions on the Laplacian. The forcing term is enforced as a constant value $f(x, y) \equiv f_0$. Such constructions are prototypical models for many physical systems, such as permeability in subsurface flows and material microstructures in elasticity. The solution term $u(x, y)$ is obtained by using a second-order finite difference scheme on a $421 \times 421$ regular grid, and then downsampled to a resolution of $85 \times 85$ for the main experiments. Following the standard practice, we use $1000$ samples for training, $200$ samples for testing, and different samples will contain different medium structures. 

\paragraph{Navier-Stokes~\cite{li21-iclr-fno}} The 2D Navier-Stokes equation models the flow of a viscous, incompressible fluid in vorticity form on the unit torus. It is mathematically formulated as: 

\begin{equation}
    \left\{
    \begin{aligned}
        \pdv{t}w(x, y, t) + u(x, y, t) \cdot \nabla w(x, y, t) &= \nu \nabla^2 w(x, y, t) + f(x, y), & (x, y) \in (0, 1)^2, \quad & t \in (0, T] \\
        \nabla \cdot u(x, y, t) &= 0, & (x, y) \in (0, 1)^2, \quad & t \in (0, T] \\ 
        w(x, y, 0) &= w_0(x, y), & (x, y) \in (0, 1)^2 \quad & \\
    \end{aligned}
    \right.
    \label{equ:2dmano:naiver-strokes}
\end{equation}

where $u(x, y, t)$ represents the velocity field, $w(x, y, t) = \nabla \times u(x, y, t)$ is the vorticity, $w_0(x, y)$ is the initial vorticity, $\nu$ is the viscosity coefficient, and $f(x, y)$ is the forcing function. In this benchmark, the viscosity $\nu$ is fixed at $10^{-5}$, and the 2D field has a resolution of $64 \times 64$. Each sample within the dataset comprises $20$ consecutive frames. The objective is to predict the subsequent ten frames based on the preceding ten. Following the common practice, we use $1000$ fluid samples with different initial conditions for training, and $200$ samples for testing.

\subsection{Baselines}

We compare GeoMaNO with 14 baselines, half of which are frequency-based NOs: FNO~\citep{li21-iclr-fno}, WMT~\citep{gupta21-nips-multiwavelet}, U-FNO~\citep{wen22-awr-u-fno}, Geo-FNO~\citep{li23-jmlr-fno-general-geometry}, U-NO~\citep{rahman23-tmlr-u-no}, F-FNO~\citep{tran23-iclr-factorized-fno}, and LSM~\citep{wu23-icml-lsm}, while the remainders are Transformer-based NOs: Galerkin~\citep{cao21-nips-galerkin}, HT-Net~\citep{liu23-corr-htnet}, OFormer~\citep{li23-tmlr-transformer-pde}, GNOT~\citep{hao23-icml-gnot}, FactFormer~\citep{li23-nips-fastformer}, ONO~\citep{xiao24-icml-ono}, and Transolver~\citep{wu24-icml-transolver}. The current SOTA is the Transolver~\citep{wu24-icml-transolver} framework. 

\vspace{5pt}
\begin{remark}[\textbf{GeoMaNO vs. Zheng et al.~\cite{zheng24-nips-aliasfreemambaneuraloperator}}]
    We have noted Zheng et al.\cite{zheng24-nips-aliasfreemambaneuraloperator}, the only peer-reviewed prior work on Mamba-based neural operators at the time of this submission. However, Zheng et al. have \textbf{\textit{not yet open-sourced their code}}, making it impossible for us to conduct experiments against them. For the theoretical completeness of this paper, we will conduct a theoretical and architectural comparison between GeoMaNO and Zheng et al.\cite{zheng24-nips-aliasfreemambaneuraloperator}. 

    Zheng et al.~\cite{zheng24-nips-aliasfreemambaneuraloperator} highlight their contributions on theoretical derivations regarding Mamba-based NO's mathematical correctness, as well as its alias-free property. Their architecture employs a U-shaped multi-resolution architecture, which is more commonly seen in vision-domain applications like segmentation and detection. 
    
    On the contrary, our novel GeoMaNO framework focuses on integrating geometric and mathematical rigor into the NO framework. We propose the GeoMamba-SSM formulation, which dampens duplicate hidden states due to the multiway cross-scan mechanism. GeoMaNO further enhances its geometric rigor by employing two-dimensional SSM recursions on 2D physical domains. We further integrate GeoMamba-SSM into the legacy NO framework. 
\end{remark}

\subsection{Evaluation Metric}

We follow the common practice founded by FNO~\cite{li21-iclr-fno} and employ the relative L2 error metric: 
\[
    \mathcal{L}(\vb{pred}, \vb{gt}) = \dfrac{\norm{\vb{pred} - \vb{gt}}_2}{\norm{\vb{gt}}_2} \ ,
\]
where $\vb{pred}$, $\vb{gt}$ separately denotes the prediction and the ground truth, and $\norm{\vb{x}}_2$ denotes the vector L2 norm. 

\subsection{Implementation Details}

\paragraph{Implementation of Baselines} All of these baselines have been widely examined in previous papers. For FNO~\cite{li21-iclr-fno} and
Geo-FNO~\cite{li23-jmlr-fno-general-geometry}, we report their results following their official papers. Other spectral baselines, i.e., WMT~\citep{gupta21-nips-multiwavelet}, U-FNO~\citep{wen22-awr-u-fno}, U-NO~\citep{rahman23-tmlr-u-no}, F-FNO~\citep{tran23-iclr-factorized-fno}, and LSM~\citep{wu23-icml-lsm}, we follow the configurations and results reported in LSM~\cite{wu23-icml-lsm}. For Transformer-based baselines, i.e., Galerkin~\citep{cao21-nips-galerkin}, HT-Net~\citep{liu23-corr-htnet}, OFormer~\citep{li23-tmlr-transformer-pde}, GNOT~\citep{hao23-icml-gnot}, FactFormer~\citep{li23-nips-fastformer}, and ONO~\citep{xiao24-icml-ono}, we follow the configurations and results reported in Transolver~\cite{wu24-icml-transolver}. For Transolver~\citep{wu24-icml-transolver}, in Table~\ref{tab:mainres_standard}, we compare with results reported in its original paper; for other experiments, we run their official code in our local environment, following its original configurations. 

\paragraph{Implementation of GeoMaNO} The training configurations are listed in Table~\ref{tab:training_detail}, and the model hyperparameters are listed in Table~\ref{tab:model_detail}. 

\begin{table*}[ht]
	\caption{Training configurations of GeoMaNO. Training configurations are directly from previous works without special tuning~\cite{wu24-icml-transolver}. For the Darcy flow benchmark, following the practice of ONO~\cite{xiao24-icml-ono}, we employ an additional spatial gradient regularization term $\mathcal{L}_{\mathrm{g}}$, in addition to the regular relative L2 loss $\mathcal{L}_{\mathrm{rL2}}$.}
	\label{tab:training_detail}
	\centering
	\small
    \begin{tabular}{l|cccccc}
        \toprule
        Benchmarks & Loss & Epochs & Initial LR & Optimizer & Scheduler & Batch Size \\
        \midrule
        
        Navier-Stokes & 
        $\mathcal{L}_{\mathrm{rL2}}$ & 
        \multirow{2}{*}{500} & 
        \multirow{2}{*}{$3 \times 10^{-4}$} & 
        \multirow{2}{*}{AdamW~\cite{loshchilov18-iclr-adamw}} & 
        \multirow{2}{*}{OneCycleLR~\cite{smith19-aiml-onecyclelr}} & 
        2 \\
        
        Darcy Flow & 
        $\mathcal{L}_{\mathrm{rL2}} + 0.1\mathcal{L}_{\mathrm{g}}$ & 
        & 
        & 
        & 
        &
        4 \\
        \bottomrule
    \end{tabular}
\end{table*}

\begin{table*}[ht]
	\caption{Model configurations of GeoMaNO.}
	\label{tab:model_detail}
	\centering
	\small
    \begin{tabular}{l|ccc}
        \toprule
        Benchmarks & Model Depth & SSM dstates & Embedding Dimensions \\
        \midrule
        
        Navier-Stokes & 
        \multirow{2}{*}{8} & 
        \multirow{2}{*}{16} & 
        256 \\
        
        Darcy Flow & 
        &
        &
        64 \\
        \bottomrule
    \end{tabular}
\end{table*}

\subsection{Error Bounds}
\label{sec:appendix:system-setup:error-bounds}

For the mathematical completeness of this paper, we report the error bounds (standard deviations) for our numerical experiments. The results are available in Table~\ref{tab:mainres_standard-stddev} and Table~\ref{tab:efficiency-stddev}. 

\begin{table*}[ht]
    \centering
    \caption{Error bounds (standard deviations) of Table~\ref{tab:mainres_standard}.}
    \label{tab:mainres_standard-stddev}
    \begin{tabular}{l|cc}
        \toprule
            \multirow{3}{*}{Model} & \multicolumn{2}{c}{Relative L2 ($\downarrow$)} \\
            \cmidrule(lr){2-3} & \ N.-S.\quad & Darcy Flow \\
        \midrule
            \textbf{GeoMaNO} & 
            \textbf{0.0370} $\pm (3.40 \times 10^{-3})$ & 
            \textbf{0.0036} $\pm (1.17 \times 10^{-4})$ \\
        \bottomrule
    \end{tabular}
\end{table*}

\begin{table*}[ht]
    \small
    \caption{Error bounds (standard deviations) of Table~\ref{tab:efficiency}.}
    \label{tab:efficiency-stddev}
    \centering

    \begin{subtable}{\textwidth}
        \centering
        \subcaption{Darcy Flow}
        \vspace{-3pt}
        
        \begin{tabular}{c|c|c|c}
            \toprule
                Model & $T_{\mathrm{train}}$ ($\downarrow$)  & $T_{\mathrm{infer}}$ ($\downarrow$)  & Mem. ($\downarrow$) \\
            \midrule
                Transolver & 
                30.99 $\pm (4.71 \times 10^{-3})$ & 
                2.33 $\pm (2.44 \times 10^{-4})$ & 
                42.55 $\pm 0.00$ \\
            \midrule
                \textbf{GeoMaNO} & 
                \textbf{23.98} $\pm (1.09 \times 10^{-3})$ & 
                \textbf{1.17} $\pm (1.98 \times 10^{-4})$ & 
                \textbf{5.21} $\pm 0.00$ \\
            \bottomrule
        \end{tabular}
    \end{subtable}

    \vspace{5pt}

    \begin{subtable}{\textwidth}
        \centering
        \subcaption{Navier-Stokes}
        \vspace{-3pt}
        
        \begin{tabular}{c|c|c|c}
            \toprule
                Model & $T_{\mathrm{train}}$ ($\downarrow$)  & $T_{\mathrm{infer}}$ ($\downarrow$)  & Mem. ($\downarrow$) \\
            \midrule
                Transolver & 
                305.33 $\pm (4.41 \times 10^{-1})$ & 
                23.47 $\pm (7.72 \times 10^{-2})$ & 
                \textbf{5.21} $\pm 0.00$ \\
            \midrule
                \textbf{GeoMaNO} & 
                \textbf{268.68} $\pm (4.71 \times 10^{-3})$ & 
                \textbf{16.61} $\pm (8.88 \times 10^{-1})$ & 
                \textbf{70.44} $\pm 0.00$ \\
            \bottomrule
        \end{tabular}
    \end{subtable}
\end{table*}

\section{Mathematical Insights on Mamba-Based Neural Operators}
\label{sec:appendix:mathematical-insights-mamba-nos}

For the theoretical completeness of this paper, we outline the mathematical derivations of Mamba-SSM's equivalence to the neural operator (NO)'s critical component, the \textit{kernel integral operator}. This section is credited to Zheng et al.~\cite{zheng24-nips-aliasfreemambaneuraloperator}. 

For more mathematical details and insights, we refer the readers to Kovachki et al.~\cite{Kovachki23-jmlr-no} for the neural operator (NO) framework's mathematical correctness and convergence guarantees, and to Zheng et al.~\cite{zheng24-nips-aliasfreemambaneuraloperator} for Mamba-SSM's equivalence to the kernel integral operators in NOs. 

\paragraph{Integral Kernels for Neural Operators} The general formulation of an NO's kernel integral operator is defined as:
\[
    \mathcal{K}_t :
    \qty{ v_t : D \to \mathbb{R}^{d_t} } \to 
    \qty{ v_{t+1} : D \to \mathbb{R}^{d_{t+1}} },
\]
which could be parameterized by a factor $\alpha$ such as:
\begin{equation}
    (\mathcal{K}_t (v_t) ; \alpha)(x) = 
    \int_D K_t(x, y, v_t(x), v_t(y)) v_t(y) \dd{y}, \quad
    \forall (x, y) \in D,
    \label{equ:no-int-kernel-with-t}
\end{equation}
where the parameter of the \textit{integral kernel} $K_t$ is learned from the training data. For example, FNO~\cite{li21-iclr-fno} employs convolution as the integral kernel, while Transformer-based NOs manifest the attention blocks~\cite{vaswani17-nips-transformer} as integral kernels. 

\paragraph{Mamba Integration} For simplicity, we omit the subscript $t$ in the right-hand-side of Eq.~\ref{equ:no-int-kernel-with-t}, which is used to denote the number of iterations during the integration flow. Eq.~\ref{equ:no-int-kernel} is reformulated as: 
\begin{equation}
    (\mathcal{K}_t (v_t) ; \alpha)(x) = 
    \int_D K(x, y, v(x), v(y)) v(y) \dd{y}, \quad
    \forall (x, y) \in D.
    \label{equ:no-int-kernel}
\end{equation}
Following the tradition of FNO~\cite{li21-iclr-fno}, we first assume that $K_t : \mathbb{R}^D \times \mathbb{R}^D \to \mathbb{R}^{d_{t+1} \times d_t}$ concerns little on the spatial variables $(v(x), v(y))$, but only on the input pair $(x, y)$. We further let 
\begin{equation}
    K_t(x, y) = C e^{Ax} \, B e^{-Ay},
    \label{equ:ktxy_ceax}
\end{equation}
where $A$, $B$ and $C$ are treated as constants for simplicity. To further ensure the possible employment of Mamba's scanning pattern~\cite{gu23-colm-mamba, gu24-icml-mamba2}, we set the integration interval to $y \in (-\infty, x)$ instead of the entire definition domain $D$. Eq.~\ref{equ:no-int-kernel} hence becomes
\begin{equation}
    (\mathcal{K}_t (v_t) ; \alpha)(x) = 
    \int_{-\infty}^x C e^{Ax} \, B e^{-Ay} \dd{y}.
    \label{equ:ktvt_ceax}
\end{equation}
Since $Ce^{Ax}$ is independent of the integral variable $y$, we could further rewrite Eq.~\ref{equ:ktvt_ceax} as:
\begin{equation}
    \left\{
    \begin{aligned}
        (\mathcal{K}_t (v_t) ; \alpha)(x) &= C h(x), \\
        h(x) &= e^{Ax} \int_{-\infty}^x B(e^{-Ay}) v(y) \dd{y}.
        \label{equ:afmano-3-10}
    \end{aligned}
    \right.
\end{equation}
By differentiating both sides of Eq.~\ref{equ:afmano-3-10} with respect to $x$, we get:
\begin{equation}
    \begin{aligned}
        h'(x) 
        &= 
        Ae^{Ax} \int_0^x B(e^{-Ay}) v(y) \dd{y} + e^{Ax} Be^{-Ax} v(x) \\
        &= 
        Ae^{Ax} \int_0^x B(e^{-Ay}) v(y) \dd{y} + Bv(x) \\
        &= 
        A h(x) + B v(x).
        \label{equ:afmano-3-11}
    \end{aligned}
\end{equation}
Combining Eq.~\ref{equ:afmano-3-10} with Eq.~\ref{equ:afmano-3-11} yields
\begin{equation}
    \left\{
    \begin{aligned}
        h'(x) &= A h(x) + B v(x), \\
        u(x) &= C h(x), 
    \end{aligned}
    \right.
    \label{equ:afmano-3-12}
\end{equation}
where $u(x) = (\mathcal{K}_t(v_t); \alpha)(x)$. Eq.~\ref{equ:afmano-3-12} directly parallels Eq.~\ref{equ:2dmano-selective-sse} in the main paper. This seamlessly marries a NO's kernel integral with a state space model (SSM)~\cite{gu21-iclr-efficiently}. Drawing inspiration from the theory of continuous systems, the goal of Eq.~\ref{equ:afmano-3-12} is to map a two-dimensional function, denoted as $v(x)$, to $u(x)$ through the hidden space $h(x)$. Here, $A$ serves as the evolution parameter, while $B$ and $C$ act as the projection parameters.

\paragraph{Discretization of Mamba Integration} To integrate Eq.~\ref{equ:afmano-3-12} into the deep learning paradigm, a discretization process is necessary. Following the Scharfetter-Gummel method~\cite{kulikovsky95-jcp-gummel}, which approximates the matrix exponential using Bernoulli polynomials, the parameters could be discretized as follows:
\begin{equation}
    \left\{
    \begin{aligned}
        \overline{\vb{A}} &= \exp(\Delta A), \\
        \overline{\vb{B}} &= (\Delta A)^{-1} (\exp(\Delta A) - \vb{I}) \ \Delta B,
    \end{aligned}
    \right.
    \label{equ:afmano-3-13}
\end{equation}
where $\Delta$ is a timescale parameter converting continuous parameters $A$ and $B$ into their discrete
counterparts $\overline{\vb{A}}$ and $\overline{\vb{B}}$. The discrete representation of Eq.~\ref{equ:afmano-3-12} can be formulated as
\begin{equation}
    \left\{
    \begin{aligned}
        h(x_k) &= \overline{\vb{A}} h(x_{k-1}) + \overline{\vb{B}} v(x_k), \\
        u(x_k) &= C h(x_k).
    \end{aligned}
    \right.
    \label{equ:afmano-3-14}
\end{equation}
Eq.~\ref{equ:afmano-3-14} directly parallels Mamba's selective state-space representation, i.e., Eq.~\ref{equ:2dmano-time-invariant-sse-discretized} in the main paper. More details can be found in the supplementary materials of Zheng et al.~\cite{zheng24-nips-aliasfreemambaneuraloperator}.

\section{Theoretical Insights on 2DMamba-SSM}
\label{sec:appendix:theoretical-insights-2dmamba}

This section discusses the 2DMamba~\cite{zhang25-cvpr-2dmamba} formulation and its mathematical and geometrical rigor. This illustrates our design choice to employ 2DMamba-SSM in our GeoMamba formulation for two-dimensional physical domains. 

\subsection{2DMamba-SSM Formulation}

In contrast to vanilla 1D Mamba-SSM, which aggregates information from a flattened 1D sequence, a 2DMamba-SSM aggregates both geometric and semantic information directly from a 2D feature map. For simplicity, we \textit{omit} the independent state dimension superscript $s$ in Eq.~\ref{equ:2d-selective-sse-discretized}, and get: 

\begin{equation}
    \left\{
    \begin{aligned}
        \vb{g}[i, j] & = \overline{\vb{A}}[i, j] \ \vb{g}[i, j - 1] + \overline{\vb{B}}[i, j] \ \vb{x}[i, j] 
        \ , \\
        \vb{h}[i, j] & = \overline{\vb{A}}[i, j] \ \vb{h}[i - 1, j] + \vb{g}[i, j]
        \ , \\
        \vb{y}[i, j] & = \sum \overline{\vb{C}}[i, j] \ \vb{h}[i, j]
        \ . 
    \end{aligned}
    \right.
    \label{equ:appendix-2d-selective-sse-discretized}
\end{equation}

Note that, for the first line in Eq.~\ref{equ:appendix-2d-selective-sse-discretized}, we define $\vb{g}[i, 0] = 0$, and thus $\vb{g}[i, 1] = \overline{\vb{B}}[i, 1] \ \vb{x}[i, 1]$. The two parameters $\overline{\vb{A}}[i, j]$ and $\overline{\vb{B}}[i, j]$ depend on the input $\vb{x}[i, j]$, regulating the information of previous hidden states $\vb{h}[i, j - 1]$, $\vb{h}[i - 1, j - 1]$, and $\vb{h}[i - 1, j - 1]$, as well as the current input $\vb{x}[i, j]$. Similarly, for the second line, we define $\vb{h}[0, i] = 0$, and thus $\vb{h}[1, j] = \vb{g}[1, j]$. 

For simplicity, in the following reasoning, we replace the square brackets with subscripts, omit the subscripts of $\overline{\vb{A}}$ and $\overline{\vb{B}}$, and expand Eq.~\ref{equ:appendix-2d-selective-sse-discretized}. With these simplications, the hidden state $\vb{h}_{i,j}$ can be formulated as the following recurrence: 

\begin{equation}
    \vb{h}_{i,j} = \sum_{i^{'} \leq i} \sum_{j^{'} \leq j} \overline{\vb{A}}^{(i - i^{'} + j - j')} \overline{\vb{B}} x_{i^{'}, j^{'}}
    \ ,
    \label{equ:2d_recursion}
\end{equation}

where $(i - i^{'} + j - j^{'})$ equals the \textbf{\textit{Manhattan distance}}~\footnote{In contrast, for the vanilla Mamba-SSM, the power of $\overline{\vb{A}}$ will be the \textbf{\textit{row-major linear distance}} between the two patches, as shown in Eq.~\ref{equ:appendix:1d-mamba-recursion-power-of-a}. This reflects the issue of geometric inconsistency in a mathematical manner.}~between patches $(i^{'}, j^{'})$ and $(i, j)$. It represents a path from $(i^{'}, j^{'})$ to $(i, j)$, which goes right horizontally and then down vertically. After two scans, the output $\vb{y}$ is aggregated from $\vb{h}$ by the parameter $\vb{C}$ similar to 1D-Mamba: $\vb{y}_{i, j} = \vb{C} \,  \vb{h}_{i, j}$. For each location $(i, j)$, the aggregation information is obtained from its upper-left locations.

\subsection{Geometric Consistency}
\label{cha:2dmamba:sec:ssm-arch:spatial-discrepancy}

\begin{figure*}[ht]
    \centering
    \includegraphics[width=0.8\linewidth]{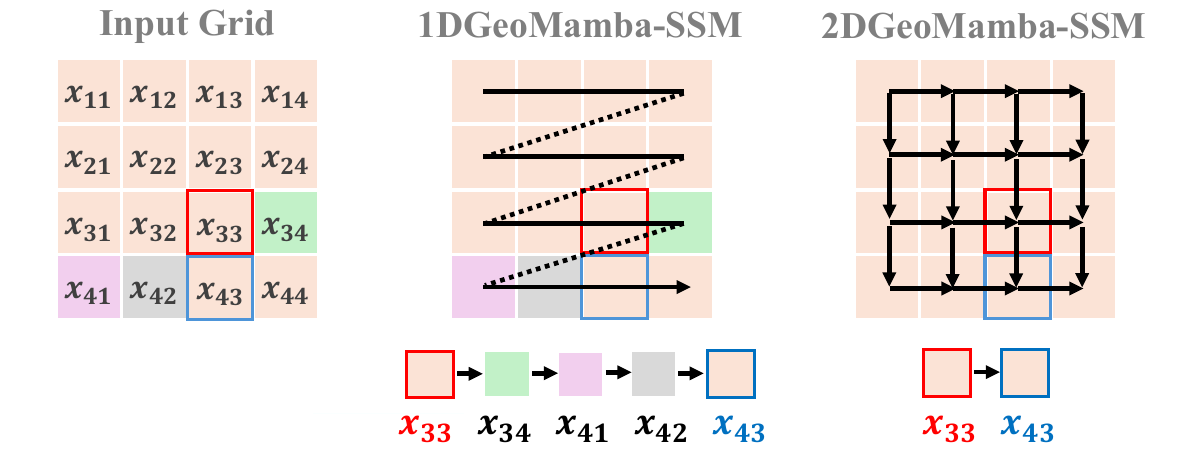}
    \caption[Scanning paths and spatial discrepancy.]{Illustration of 1D v.s. 2D scanning paths and spatial discrepancy. \textbf{Left}: The two-dimensional input grid. \textbf{Middle}: 1D Mamba-SSM flattens the input into a 1D sequence, but adjacent patches (the {\color{red} red} and {\color{blue} blue}) are far away in this sequence, leading to the \textbf{\textit{geometric inconsistency}} in the hidden states. \textbf{Right}: 2DMamba-SSM the input grid in a 2D manner, and maintains spatial continuity.}
    \label{fig:spatial-discrepancy}
\end{figure*}

In this subsection, we will use a concrete example to illustrate our GeoMamba-SSM's geometric rigor. Fig.~\ref{fig:spatial-discrepancy} illustrates the 1D scanning path employed by the vanilla Mamba-SSM vs. the 2D scanning path employed by our 2DGeoMamba-SSM. 

As illustrated in Fig.~\ref{fig:spatial-discrepancy}, the hidden state of the vanilla Mamba~\cite{gu23-colm-mamba} on a flattened image is 

\begin{equation}
    \vb{h}^{\mathrm{1D}}_{i} = \sum_{i'\leq i} \overline{\vb{A}}^{(i-i')} \overline{\vb{B}} \, \vb{x}_{i'}
    \, ,
    \label{equ:appendix:1d-mamba-recursion-power-of-a}
\end{equation}

where $i$ is an 1D index. The order $i - i'$ is the distance between $i$ and $i'$ in a flattened sequence (a higher order results in forgetting). From Fig.~\ref{fig:spatial-discrepancy}, flattening in the 1D direction results in a loss of spatial structure in other directions, and thus is sub-optimal, regardless of various formulations for flattening. We denote this phenomenon as \textbf{\textit{spatial discrepancy}}. 

To solve this problem, 2DMamba-SSM formulates Eq.~\ref{equ:2d_recursion}. We will then use a concrete example to compare these two formulations. 

For an input feature map of size $3 \times 3$, the last hidden state of the vanilla Mamba-SSM is formulated as:

\begin{equation}
    \begin{aligned}
        \vb{h}^{\mathrm{1D}}_{3, 3} = 
        &
        \overline{\vb{A}}^{\color{red} 8} \vb{x}_{1, 1} + 
        \overline{\vb{A}}^{\color{red} 7} \vb{x}_{1, 2} + 
        \overline{\vb{A}}^{\color{red} 6} \vb{x}_{1, 3} + 
        \\ 
        &
        \overline{\vb{A}}^{\color{red} 5} \vb{x}_{2, 1} + 
        \overline{\vb{A}}^{\color{red} 4} \vb{x}_{2, 2} + 
        \overline{\vb{A}}^{\color{red} 3} \vb{x}_{2, 3} + 
        \\
        &
        \overline{\vb{A}}^{\color{red} 2} \vb{x}_{3, 1} + 
        \overline{\vb{A}}^{\color{red} 1} \vb{x}_{3, 2} + 
        \overline{\vb{A}}^{\color{red} 0} \vb{x}_{3, 3} 
        \ ,
    \end{aligned}
\end{equation}

whereas 2DMamba-SSM formulation yields: 

\begin{equation}
    \begin{aligned}        
        \vb{h}^{\mathrm{2D}}_{3, 3} = 
        &
        \overline{\vb{A}}^{\color{red} 4} \vb{x}_{1, 1} + 
        \overline{\vb{A}}^{\color{red} 3} \vb{x}_{1, 2} + 
        \overline{\vb{A}}^{\color{red} 2} \vb{x}_{1, 3} + 
        \\ 
        &
        \overline{\vb{A}}^{\color{red} 3} \vb{x}_{2, 1} + 
        \overline{\vb{A}}^{\color{red} 2} \vb{x}_{2, 2} + 
        \overline{\vb{A}}^{\color{red} 1} \vb{x}_{2, 3} + 
        \\
        &
        \overline{\vb{A}}^{\color{red} 2} \vb{x}_{3, 1} + 
        \overline{\vb{A}}^{\color{red} 1} \vb{x}_{3, 2} + 
        \overline{\vb{A}}^{\color{red} 0} \vb{x}_{3, 3} 
        \ .
    \end{aligned}
\end{equation}

It can be observed that, for the vanilla Mamba-SSM, the coefficient of $\vb{x}_{1, 1}$ is $\overline{\vb{A}}^{\color{red} 8}$, where the power reflects the \textbf{\textit{row-major linear distance}} between patches $(1, 1)$ and $(3, 3)$. On the contrary, 2DMamba-SSM's coefficient for $\vb{x}_{1, 1}$ is $\overline{\vb{A}}^{\color{red} 4}$, where the power reflects the \textbf{\textit{Manhattan distance}} between patches $(1, 1)$ and $(3, 3)$. Mathematically speaking, 1D Mamba's higher order of $\overline{\vb{A}}$ leads to more forgetting and loss of 2D structure. Meanwhile, semantically speaking, given that these adjacent patches are spatially connected, our 2DMamba-SSM better reflects their neighboring relationship. On the contrary, the 1D Mamba creates spatial discrepancy and may dilute the information from vertically near patches.

\section{Hardware-Aware Implementation of GeoMamba-SSM}
\label{sec:appendix:hardware-geomamba}

Our GeoMamba-SSM formulation is a novel neural architecture, and it comes with no ready-to-use implementation~\footnote{The vanilla Mamba-SSM has an efficient GPU implementation, yet it does not support our geometric correction components.}~. As a remedy, we propose a novel GPU algorithm for our GeoMamba-SSM. 

GPU algorithms must be designed with special care for the GPU memory hierarchy, minimizing memory transfers between on-chip registers/caches and off-chip GPU VRAM (a.k.a. HBM). In GPU algorithms, data is transferred from HBM to on-chip memory for computation, and the results are stored back to HBM to vacate on-chip memory for succeeding on-chip operations. Memory transfers are expensive. Therefore, instead of computation, many GPU algorithms, including the vanilla 1D Mamba-SSM~\cite{gu23-colm-mamba, gu24-icml-mamba2}, are memory-bounded. Here, a naive implementation will ruin the GPU memory hierarchy and thus suffer from prohibitively low throughput and large GPU VRAM requirements.

In this section, we first revisit the GPU memory hierarchy and the hardware-aware 1D selective SSE solver proposed by Mamba \cite{gu23-colm-mamba}. Next, we analyze the major challenges for implementing our GeoMamba-SSM kernels. Then, we present our selective scanning algorithm and the GPU algorithm to implement GeoMamba-SSM in detail. 

\begin{figure}[ht]
    \centering
    \includegraphics[width=0.6\linewidth]{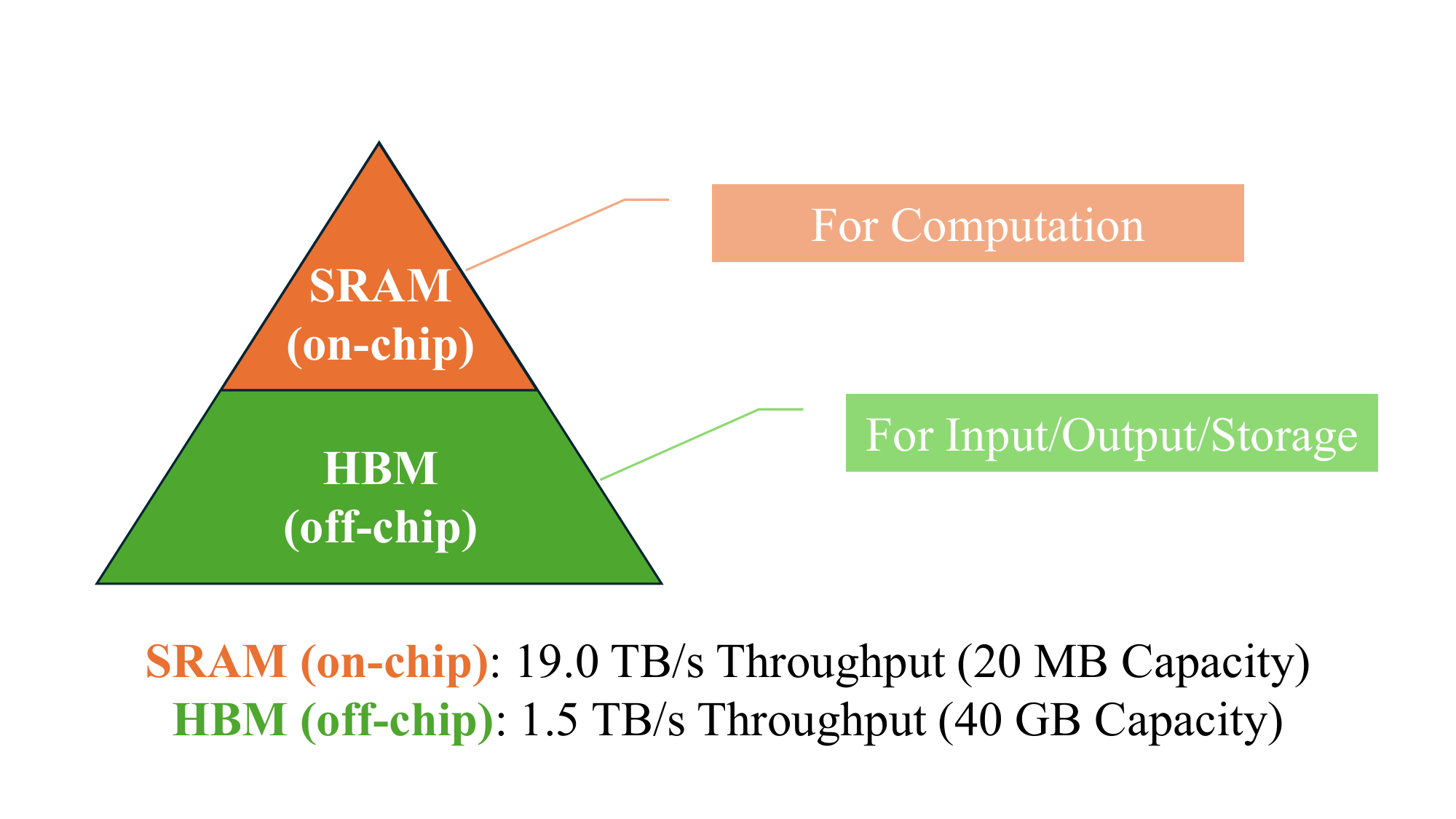}
    \caption[GPU memory hierarchy.]{Illustration of the typical GPU Memory hierarchy. The typical GPU memory hierarchy. On-chip SRAMs are small but fast; off-chip HBMs have large capacities but are slow. Data of throughput and capacity reflect an NVIDIA Quadro RTX 8000 GPU.}
    \label{fig:gpu-mem-hierachy}
\end{figure}

\paragraph{GPU Memory Hierarchy} Fig.~\ref{fig:gpu-mem-hierachy} illustrates the memory hierarchy of modern GPUs. The {\color{green} {green}} area represents off-chip GPU memory. \textbf{\textit{Off-chip memory}} is characterized by low speed and high capacity, and typically consists of High-Speed Memory (HBM). Here, for simplicity, it is referred to as HBM. On the other hand, the {\color{orange} {orange}} area denotes on-chip GPU memory. \textbf{\textit{On-chip memory}} is characterized by high speed and low capacity, and typically consists of Static Random-Access Memory (SRAM, a.k.a. caches) and registers. Again, for simplicity, it is referred to as SRAM. 

\begin{figure}[ht]
    \centering
    \includegraphics[width=0.7\linewidth]{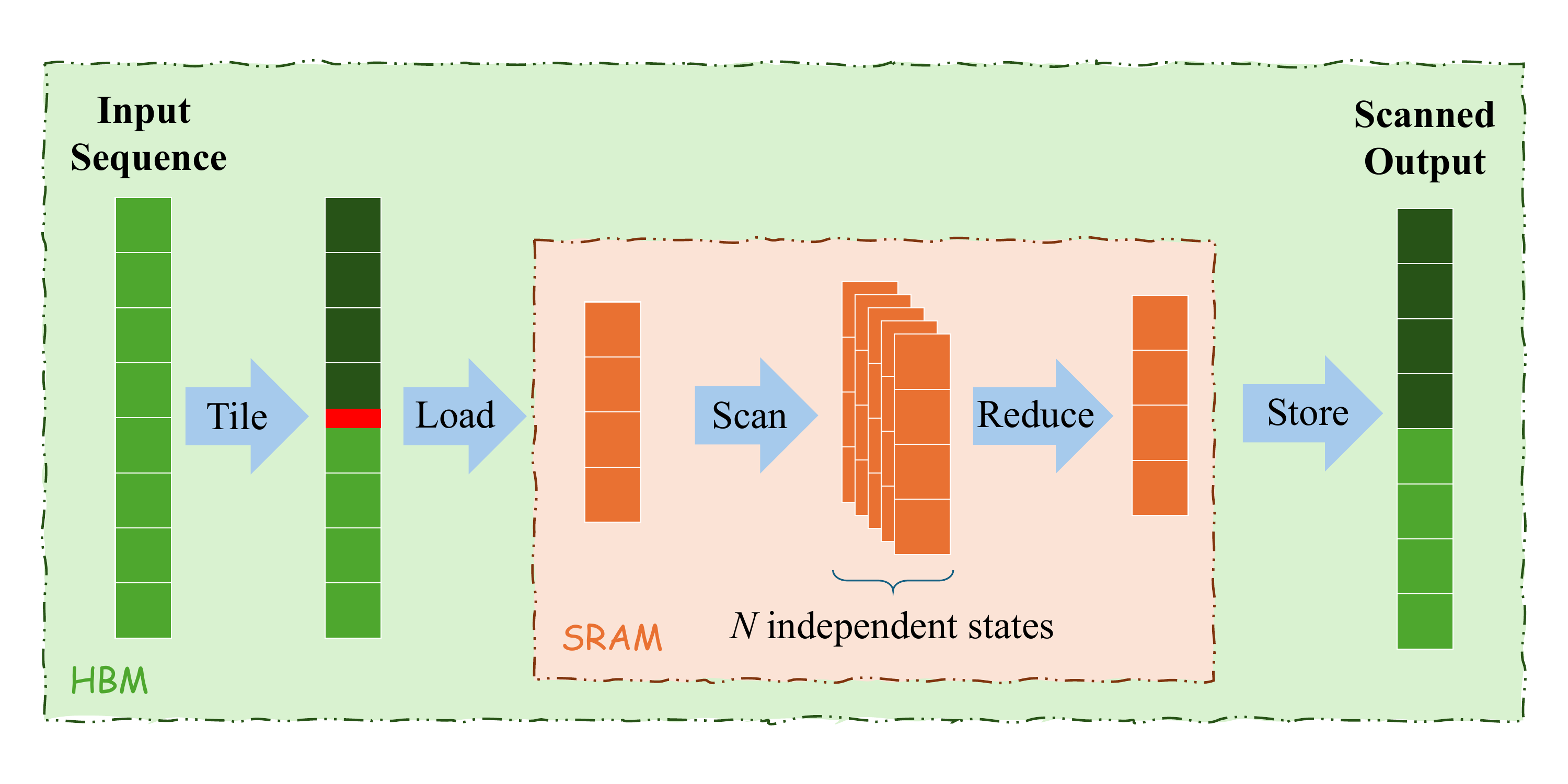}
    \caption[Mamba's 1D parallel scan.]{Illustration of  Mamba's 1D scanning operator. Here, {\color{orange} {orange}} color represents data and operations on SRAM; while  {\color{green} {green}} color represents those on HBM. Mamba's 1D scanning operator intakes a flattened sequence on HBM. It tiles the input into sub-sequences. Each sub-sequence is loaded from HBM to SRAM, scanned and reduced across $N$ intermediate mutually-independent \textit{state dimensions}, and then stored back to HBM. The total memory access complexity is $\mathcal{O}(L)$.}
    \label{fig:mamba-scan-1d}
\end{figure}

In GPU algorithms, data is transferred from HBM to SRAM for computation, and the results are stored back to HBM to vacate SRAM for succeeding computation, as illustrated in Fig.~\ref{fig:mamba-scan-1d}. Memory transfers are expensive. Therefore, instead of computation, many GPU algorithms \cite{dao22-nips-flashattention, dao23-iclr-flashattention2} are bounded by memory. Mamba-SSM~\cite{gu23-colm-mamba} is also \textit{memory-bounded}.

\paragraph{The Vanilla Mamba-SSM} The neural module of Mamba-SSM iterates through the input sequence $\vb{x}[t]$ and regresses the hidden states $\vb{h}_{s}[t]$ via a parallel prefix scan \cite{ladner80-jacm-parallel-scan, sengupta07-siggraph-gpu-scan} operation. In the literature of Computer Vision, this solving process is referred to as a \textit{selective scan} due to the selective nature of the SSE coefficients. 

Mamba-SSM is efficient because it respects the GPU memory hierarchy by 1D tiling and caching. As shown in Fig.~\ref{fig:mamba-scan-1d}, a long feature sequence in HBM is divided into smaller tiles. Each tile is loaded into SRAM, scanned across $N$ independent state dimensions, aggregated into a single output, and finally stored back to HBM. The intermediate results of the $N$ state dimensions are \textit{\textbf{not} explicitly materialized on HBM} and will be recomputed during back-propagation. The overall memory access complexity is $\mathcal{O}(L)$, where $L$ denotes the sequence length.

\paragraph{GeoMamba-SSM} Algorithm~\ref{alg:2dmano-2dmamba-forward} details our novel GeoMamba-SSM scanner module. In Algorithm~\ref{alg:2dmano-2dmamba-forward}, $\circ$ denotes element-wise multiplication. Algorithm~\ref{alg:2dmano-2dmamba-forward} is designed as a hardware-aware GPU kernel, which operates at the granularity of a single GPU thread. Our geometric correction mechanism is applied at line 24 of Algorithm~\ref{alg:2dmano-2dmamba-forward}. 

\begin{algorithm}[ht]
    \caption{Hardware-aware GeoMamba-SSM Module.}
    \label{alg:2dmano-2dmamba-forward}
    \begin{algorithmic}[1]
        \small
        \Require{The current two-dimensional patch grid $\widehat{\vb{X}}: (B, ED, H, W)$;}
        \Require{Time step $\Delta_t: (B, ED, H, W)$ and bias term $\Delta_{\mathrm{bias}}: (N,)$;}
        \Require{SSM coefficients $\widehat{\vb{A}_t}: (N, ED)$ and $\widehat{\vb{B}_t}: (B, H, W)$;}
        \Require{Aggregation weight $\widehat{\vb{C}_t}: (B, N)$; }
        \Require{Skip-connection weight $\widehat{\vb{D}_t}: (ED,)$;}
        \Require{Geometric Correction weight $\widehat{\vb{R}_t}: (N, ED)$;}
        \Require{GPU tile size $T$.}
        
        \Ensure{Aggregated patch grid $\vb{Y}: (B, ED, H, W)$}.
        
        \State $K_H, K_W = \left\lceil H / T \right\rceil, \left\lceil W / T \right\rceil$;

        \State On-chip $\Delta_{\mathrm{bias}} = $ Load from HBM;
        \State On-chip $D = $ Load from HBM;
        
        \For{$k_h = 1$ to $K_H$ and $k_w = 1$ to $K_W$} \Comment{Loop for $K_H \times K_W$ tiles.}
            \State On-chip {$\vb{x}[k_h, k_w]: (T, T) = $ Load from HBM;}
            \State On-chip {$\Delta[k_h, k_w]: (T, T) = $ Load from HBM;}
            \State $\Delta_{\vb{x}}: (T, T) = \mathrm{softplus}(\Delta \circ \vb{x}[k_h, k_w] + \Delta_{\mathrm{bias}})$;.
            \State $\vb{y}[k_h, k_w]: (T, T) = \vb{0}$;
            
            \For{Independent SSM dstates $s = 1$ to $N$}
                \State On-chip $A_s = $ Load from HBM; \Comment{Scalar.}
                \State On-chip $\vb{B}_s[k_h, k_w]: (T, T) = $ Load from HBM; \Comment{Matrix.}
                \State On-chip $C_s = $ Load from HBM; \Comment{Scalar.}
                \State On-chip $R_s = $ Load from HBM; \Comment{Scalar.}
                \State $\vb{B}^s_\Delta \vb{x}: (T, T) = \vb{B}_s \circ \Delta \circ \vb{x}[k_h, k_w]$;
                \State $\vb{A}^s_\Delta: (T, T) = \vb{A}_s[k_h, k_w] \circ \Delta_{\vb{x}}$;
                
                \State \textcolor{brown}{\# Scan each row in parallel based on $\vb{A}^s_\Delta$.}
                \State On-chip $\vb{P}^{\mathrm{hor}} = $ Load $\vb{P}^{\mathrm{hor}}[k_h, k_w - 1]$ from HBM;
                \State $\vb{g}_s: (T, T) = \mathrm{parallel\_horizontal\_scan}
                (\exp(\vb{A}^s_\Delta), 
                \vb{B}^s_\Delta \vb{x}, 
                \vb{P}^{\mathrm{hor}})$; 
                \State Store last column of $\vb{g}_s$ as $\vb{P}^{\mathrm{hor}}[k_h, k_w]$ to HBM;

                \State \textcolor{brown}{\# Scan each column in parallel based on $\vb{g}_s$.}
                \State On-chip $\vb{P}^{\mathrm{ver}} = $ Load $\vb{P}^{\mathrm{hor}}[k_h - 1, k_w]$ from HBM; 
                \State $\vb{h}_s: (T, T) = \mathrm{parallel\_vertical\_scan}(\exp(\vb{A}^s_\Delta), \vb{g}_s, \vb{P}^{\mathrm{ver}})$;
                \State Store last row of $\vb{h}_s$ as $\vb{P}^{\mathrm{ver}}[k_h, k_w]$ to HBM;
                
                \State $\vb{y}[k_h, k_w] = \vb{y}[k_h, k_w] + C_s \circ (\vb{h}_s - R_s \vb{x}[k_h, k_w])$; \Comment{Geometric correction and aggregation.}
                
            \EndFor \Comment{End state dimensions.}
            \State $\vb{y}[k_h, k_w] = \vb{y}[k_h, k_w] + D \,  \vb{x}[k_h, k_w]$. \Comment{Learnable skip-connection.}
            \State {Store $\vb{y}[k_h, k_w]$ to HBM.}
        \EndFor  \Comment{End tiles.}
    \end{algorithmic}
\end{algorithm}

The proposed hardware-aware GeoMamba-SSM module optimizes memory transactions by conducting the geometric corrections at the GPU thread level, instead of appending PyTorch-level successors in the computation graph. This eliminates unnecessary GPU kernel launches and avoids creating nodes for our geometric correction in the computation graph. Our hardware-aware implementation of GeoMamba-SSM maintains an overall $\mathcal{O}(L)$ memory complexity, avoiding a multiplication with the Mamba dstate dimension $N$. This enhances the overall computational efficiency of our GeoMaNO framework.

\section{Generalization to Other Domains}
\label{sec:appendix:generalization-to-other-domains}

To further evaluate the geometric rigor and generalization power of our GeoMamba-SSM formulation, we integrate GeoMamba-SSM in the SOTA Mamba-based foundation model, the VMamba~\cite{liu24-nips-vmamba} framework, for a standard downstream vision-domain benchmark: ImageNet-100 classification benchmark~\cite{deng09-cvpr-imagenet}. As shown in Table~\ref{tab:imagenet100-classification}, GeoMamba-SSM outperforms vanilla Mamba-SSM~\cite{gu23-colm-mamba} by $0.22\%$, which is a significant performance gain in the field of natural images. This showcases the necessity of our geometric correction mechanism.

\begin{table*}[ht]
    \small
    \caption{Results on ImageNet-100 classification benchmark. The best configuration and result is highlighted in \textbf{bold}.}
    \label{tab:imagenet100-classification}
    \centering

    \begin{tabular}{c|cc}
        \toprule
            SSM & Geo. Coeff. & Top-1 Acc. (\%) \\
        \midrule
            Mamba-SSM & None & 90.90 \\
        \midrule
            \textbf{GeoMamba-SSM} & \textbf{0011} & \textbf{91.12} \\
            GeoMamba-SSM & 0111 & 90.90 \\
        \bottomrule
    \end{tabular}
\end{table*}

\begin{figure}[H]
    \centering
    \begin{subfigure}[h]{0.3\textwidth}
        \centering
        \includegraphics[width=\linewidth]{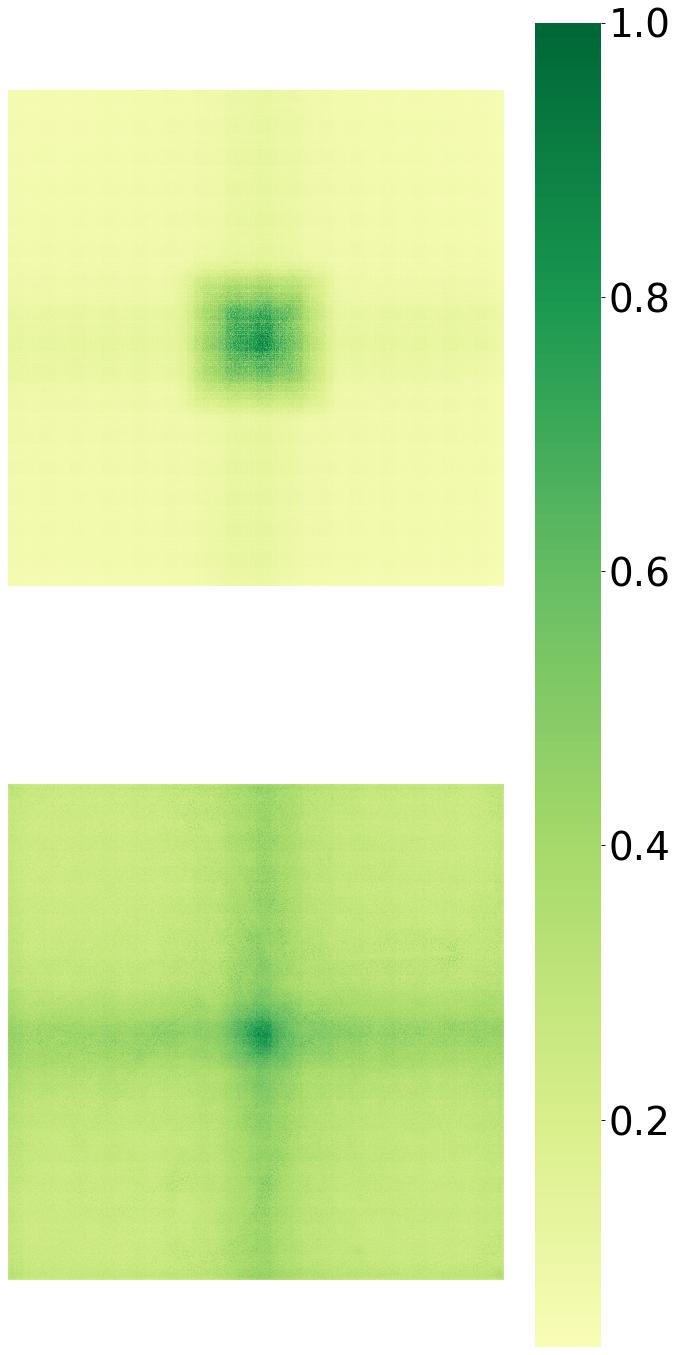}
        \caption{Vanilla Mamba}
    \end{subfigure}
    \begin{subfigure}[h]{0.3\textwidth}
        \centering
        \includegraphics[width=\linewidth]{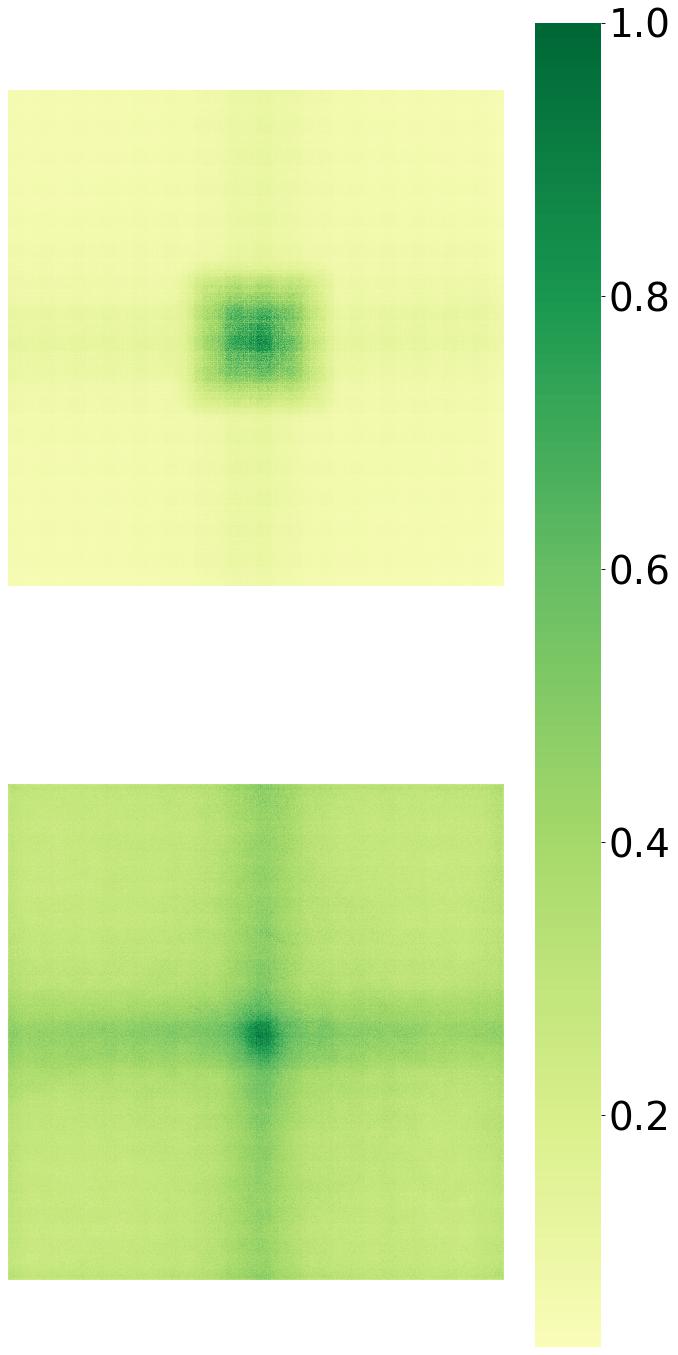}
        \caption{Geo. Coeff. 0011}
    \end{subfigure}
    \begin{subfigure}[h]{0.3\textwidth}
        \centering
        \includegraphics[width=\linewidth]{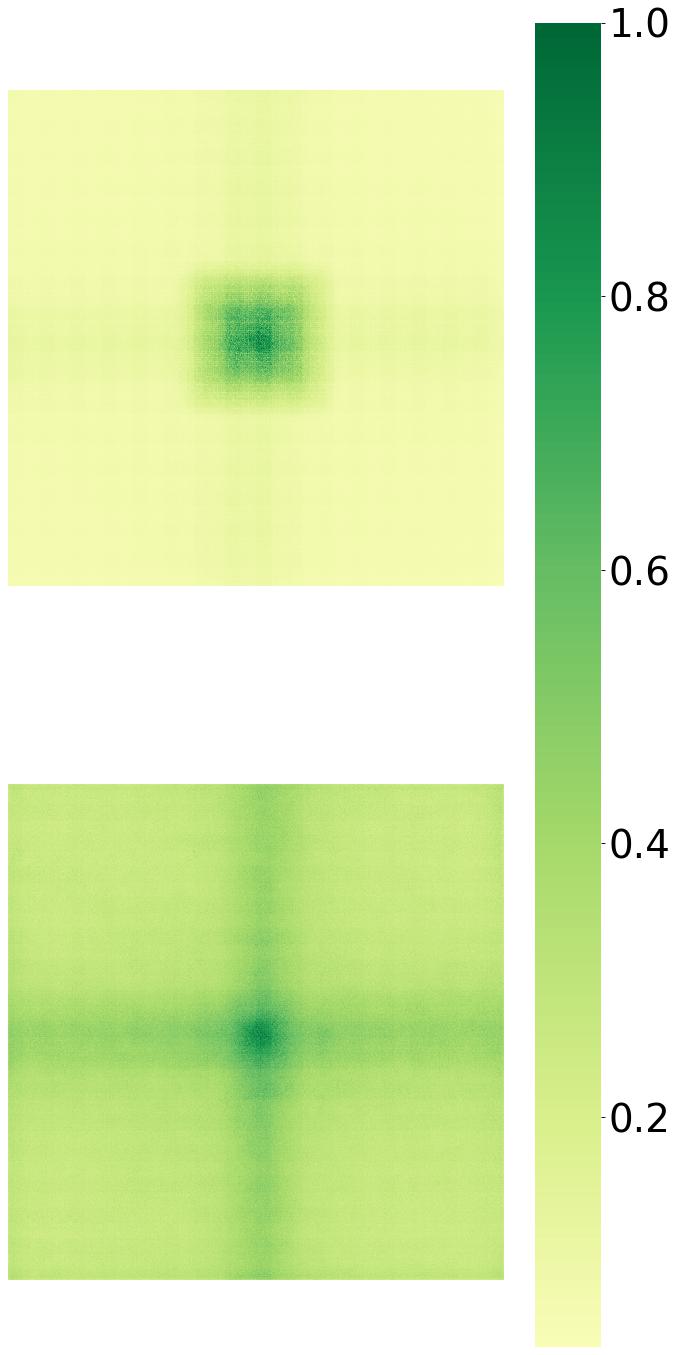}
        \caption{Geo. Coeff. 0111}
    \end{subfigure}
    \caption[Comparasion of ERFs.]{Comparasion of ERFs on ImageNet100 classification tasks. Pixels with higher intensity indicate larger responses regarding the central pixel. With our geometric coefficients, the potential space focuses less along the central cross region, showcasing GeoMamba-SSM's geometric rigor.}
    \label{fig:erf}
\end{figure}

\paragraph{Visualization of Effective Receptive Field (ERF)} ERF refers to the region in the input space that contributes to the activation of a certain output unit~\cite{luo16-nips-erf}. As illustrated in Fig.~\ref{fig:erf}, the latent space of the central pixel is less concentrated along the central cross~\footnote{This cross-like pattern is caused by the cross-scan pattern.}~, and distributes more across the whole domain. This highlights GeoMamba-SSM's geometric rigor. 

\end{document}